%% file: main.tex
\documentclass[runningheads]{llncs}
\usepackage{graphicx}
\usepackage{comment}
\usepackage{amsmath,amssymb} 
\usepackage{color}
\usepackage{bm}
\usepackage{cite}
\usepackage{mathtools}
\usepackage{microtype}
\usepackage{multirow}
\usepackage{subfigure}
\usepackage{booktabs} 
\usepackage{hyperref}
\usepackage{array}
\usepackage{url}
\usepackage{float}

\usepackage[resetlabels]{multibib}
\newcites{Supp}{References}

\usepackage{times} 

\makeatletter
\newcommand{\printfnsymbol}[1]{%
  \textsuperscript{\@fnsymbol{#1}}%
}
\makeatother

\DeclareMathOperator{\E}{\mathbb{E}}

\definecolor{Red}{rgb}{1,0,0}
\definecolor{SH}{rgb}{0.5,0.5,1}

\raggedcolumns
\begin{document}
\pagestyle{headings}
\mainmatter
\def\ECCVSubNumber{1322}  

\title{Diverse and Admissible Trajectory Forecasting through Multimodal Context Understanding}

\titlerunning{Diverse and Admissible Trajectory Forecasting}
%
\author{
Seong Hyeon Park \index{Park, Seong Hyeon} \inst{1}
\and 
Gyubok Lee\inst{2} \and 
Manoj Bhat\inst{3}\thanks{Authors contributed equally}\and 
Jimin Seo\inst{4}\printfnsymbol{1} 
\and Minseok Kang \inst{5} \and
Jonathan Francis\inst{3,6} \and 
Ashwin Jadhav\inst{3} \and
Paul Pu Liang \index{Liang, Paul Pu} \inst{3} \and \\
Louis-Philippe Morency \index{Morency, Louis-Philippe} \inst{3}} 


%
\authorrunning{S. Park, et al.}

\institute{Hanyang University, Seoul, Korea\\
\email{shpark@spa.hanyang.ac.kr} \and
Yonsei University, Seoul, Korea\\
\email{glee48@yonsei.ac.kr} \and
Carnegie Mellon University, Pittsburgh, PA, USA\\
\email{\{mbhat,jmf1,arjadhav,pliang,morency\}@cs.cmu.edu} \and
Korea University, Seoul, Korea\\
\email{jmseo0607@korea.ac.kr} \and
Sogang University, Seoul, Korea\\
\email{ahstarwab@sogang.ac.kr} \and
Bosch Research Pittsburgh, Pittsburgh, PA, USA}

\maketitle{}


\begin{abstract}
Multi-agent trajectory forecasting in autonomous driving requires an agent to accurately anticipate the behaviors of the surrounding vehicles and pedestrians, for safe and reliable decision-making. Due to partial observability in these dynamical scenes, directly obtaining the posterior distribution over future agent trajectories remains a challenging problem. In realistic embodied environments, each agent's future trajectories should be both \textit{diverse} since multiple plausible sequences of actions can be used to reach its intended goals, and \textit{admissible} since they must obey physical constraints and stay in drivable areas. In this paper, we propose a model that synthesizes multiple input signals from the multimodal world|the environment's scene context and interactions between multiple surrounding agents|to best model all diverse and admissible trajectories. We compare our model with strong baselines and ablations across two public datasets and show a significant performance improvement over previous state-of-the-art methods. Lastly, we offer new metrics incorporating admissibility criteria to further study and evaluate the diversity of predictions. Codes are at: \url{https://github.com/kami93/CMU-DATF}.

\keywords{Trajectory Forecasting \and Diversity \and Admissibility \and Generative Modeling \and Autonomous Driving}
\end{abstract}


\section{Introduction}

\begin{figure}[t]
\centering
\includegraphics[height=4cm]{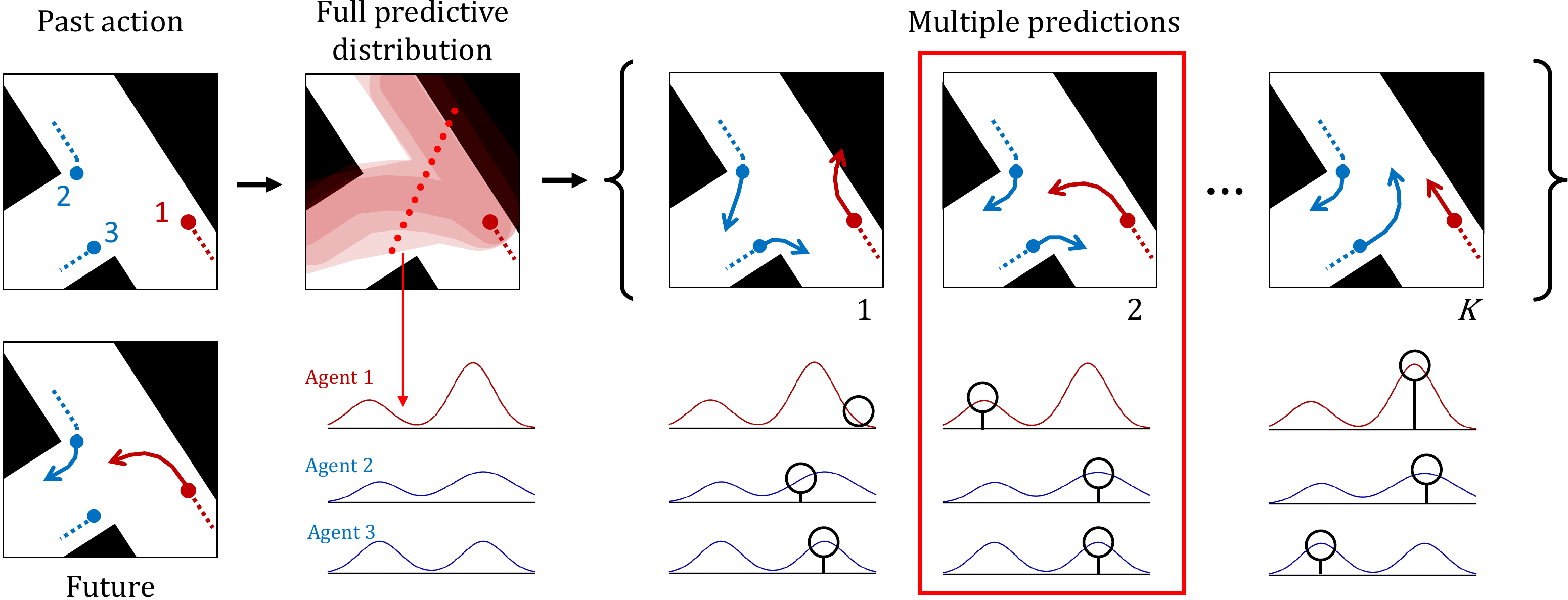}
\caption{Diverse and admissible trajectory forecasting. Based on the existing context, there can be multiple valid hypothetical future trajectories. Therefore, the predictive distribution of the trajectories should reflect various modes, representing different plausible goals (\textit{diversity}) while penalizing implausible trajectories that either conflict with the other agents or are outside valid drivable areas (\textit{admissibility}).}
\label{fig:task_definition}
\end{figure}

Trajectory forecasting is an important problem in autonomous driving scenarios, where an autonomous vehicle must anticipate the behavior of other surrounding agents (e.g., vehicles and pedestrians), within a dynamically-changing environment, in order to plan its own actions accordingly. However, since the contexts of agents' behavior such as intentions, social interactions, or environmental constraints are not directly observed, predicting future trajectories is a challenging problem~\cite{zhao2019multi, rhinehart2019precog,tang2019multiple}.
It requires an estimation of most likely agent actions based on key observable environmental features (e.g., road structures, agent interactions) as well as the simulation of agents' hypothetical future trajectories toward their intended goals. In realistic embodied environments, there are multiple plausible sequences of actions that an agent can take to reach its intended goals. However, each trajectory must obey physical constraints (e.g., Newton's laws) and stay in the statistically plausible locations in the environment (i.e., the drivable areas). In this paper, we refer to these attributes as \textit{diverse} and \textit{admissible} trajectories, and illustrate some examples in Fig. ~\ref{fig:task_definition}. Achieving diverse and admissible trajectory forecasting for autonomous driving allows each agent to make the best predictions, by taking into account all valid actions that other agents could take.

To predict a diverse set of admissible trajectories, each agent must understand its \textit{multimodal} environment, consisting of the scene context as well as interactions between other surrounding agents. Scene context refers to the typical and spatial activity of surrounding objects, presence of traversable area, etc. which contribute to forecasting next maneuvers. These help to understand some semantic constraints on the agent's motion (e.g., traffic laws, road etiquette) and can be inferred from the present and corresponding mutlimodal data i.e spatial as well as social, temporal motion behavior data. Therefore, the model's ability to extract and meaningfully represent multimodal cues is crucial.

Concurrently, another challenging aspect of trajectory forecasting lies in encouraging models to make diverse predictions about future trajectories.
Due to high-costs in data collection, 
diversity is not explicitly present in most public datasets, but only one annotated future trajectories~\cite{krajewski2018highd, chang2019argoverse, ma2019trafficpredict}. Vanilla predictive models that fit future trajectories based only on the existing annotations would severely underestimate the diversity of all possible trajectories. In addition, measuring the quality of predictions using existing annotation-based measures (e.g., displacement errors~\cite{rudenko2019human}) does not faithfully score diverse and admissible trajectory predictions.

\begin{figure}[t]
\centering
\includegraphics[width=10cm]{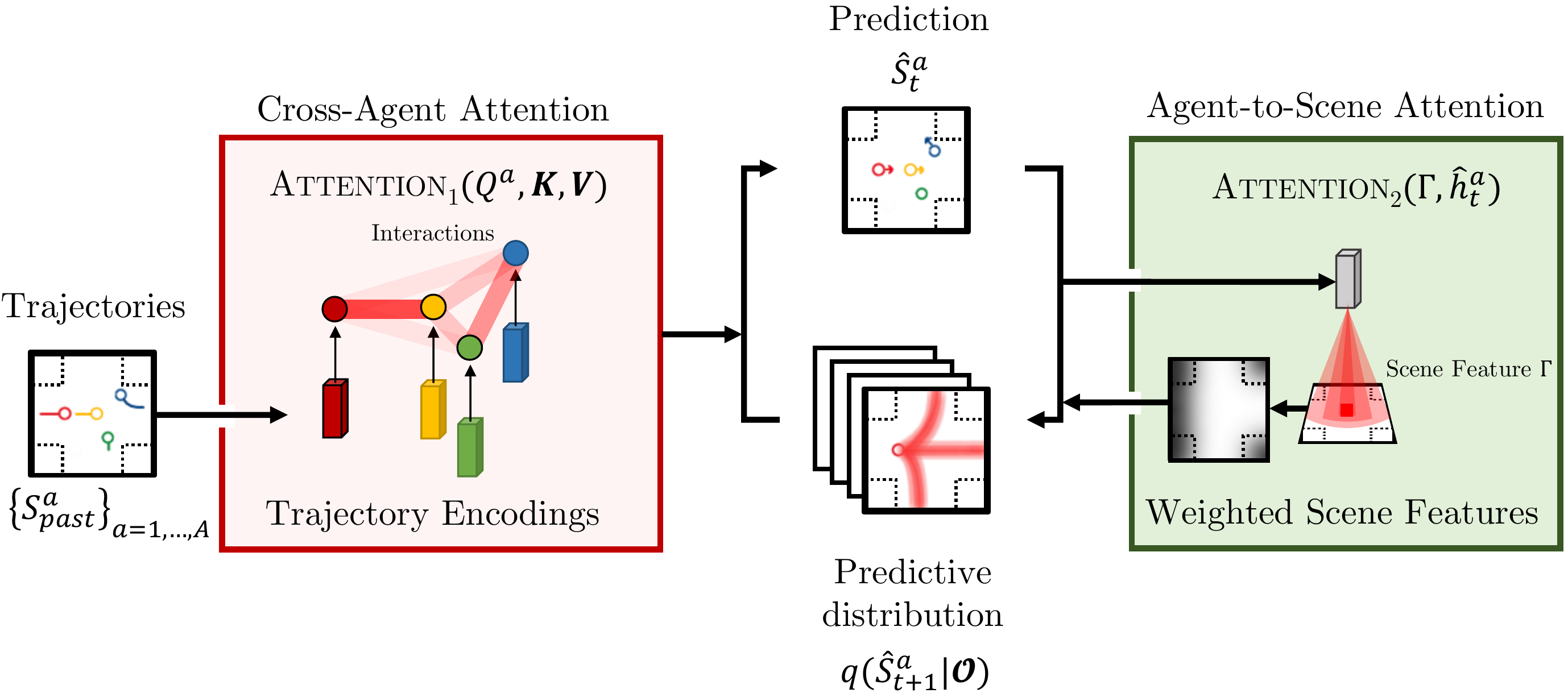}
\caption{Overview of our multimodal attention approach. Best viewed in color. The cross-agent attention module (left) generates an attention map over the encoded trajectories of neighboring agents. The agent-to-scene attention model (right) generates an attention map over the scene, based on the drivable-area map.}
\label{fig:model_diagram}
\end{figure}

As a step towards multimodal understanding for diverse and admissible trajectory forecasting, our contributions are \textit{three}-fold:
\renewcommand{\labelitemi}{$\bullet$}
\begin{enumerate}
    \item We propose a model that addresses the lack of diversity and admissibility for trajectory forecasting through the understanding of the multimodal environmental context. As illustrated in Fig.~\ref{fig:model_diagram}, our approach explicitly models agent-to-agent and agent-to-scene interactions through ``self-attention'' \cite{vaswani2017attention} among multiple agent trajectory encodings, and a conditional trajectory-aware ``visual attention'' \cite{xu2015show} over the map, respectively. Together with a constrained flow-based decoding, trained with symmetric cross-entropy~\cite{rhinehart2018r2p2}, this allows our model to generate diverse and admissible trajectory candidates by fully integrating all environmental contexts.
    \item We propose a new annotation-free approach to estimate the true trajectory distribution based on the drivable-area map. This approximation is utilized for evaluating hypothetical trajectories generated by our model during the learning process. Previous methods \cite{rhinehart2018r2p2} depend on ground-truth (GT) recordings to model the real distribution; for most of the time, only one annotation is available per agent. Our approximation method does not rely on GT samples and empirically facilitates greater diversity in the predicted trajectories while ensuring admissibility.
    \item We propose a new metric, Drivable Area Occupancy (DAO), to evaluate the diversity of the trajectory predictions while ensuring admissibility. This new metric makes another use of the drivable-area map, without requiring multiple annotations of trajectory futures. We couple this new metric with standard metrics from prior art, such as Average Displacement Error (ADE) and Final Displacement Error (FDE), to compare our model with existing baselines.
\end{enumerate}
Additionally, we publish tools to replicate our data and results which we hope will advance the study of diverse trajectory forecasting. Our codes are available at: \url{https://github.com/kami93/CMU-DATF}.

\section{Related Work}
\label{sec:related_work}

\noindent \textbf{Multimodal trajectory forecasting} requires a detailed understanding of the agent's environment. Many works integrate information from multiple sensory cues, such as LiDAR point-cloud information to model the surrounding environment \cite{lee2017desire, rhinehart2018r2p2, rhinehart2019precog}, high dimensional map data to model vehicle lane segmentation\cite{zhao2019multi, bansal2018chauffeurnet, casas2018intentnet}, and RGB image to capture the environmental context \cite{lee2017desire, sadeghian2019sophie, ma2019trafficpredict}. Other methods additionally fuse different combinations of interactions with the intention of jointly capturing all interactions between the agents  \cite{alahi2016social, gupta2018social, sadeghian2019sophie, ma2019trafficpredict}. Without mechanisms to jointly and explicitly model such agent-to-scene and agent-to-agent relations, we hypothesize that these models are unable to capture complex interactions in the high-dimensional input space and propose methods to explicitly model these interactions via sophisticated attention mechanisms.

\noindent \textbf{Multi-agent modeling} aims to learn representations that summarize the behavior of one agent given its surrounding agents. These interactions are often modeled through either spatial-oriented methods or neural attention-based methods. Some of the spatial-oriented methods simply take into account agent-wise distances through a relative coordinate system~\cite{rhinehart2019precog, park2018sequence, kim2017probabilistic, bansal2018chauffeurnet}, while others utilize sophisticated pooling approaches across individual agent representations~\cite{lee2017desire, gupta2018social, deo2018convolutional, zhao2019multi}. On the other hand, the attention-based methods use the attention architecture to model multi-agent interaction in a variety of domains including pedestrians \cite{vemula2018social, fernando2018soft+, sadeghian2019sophie} and vehicles \cite{tang2019multiple, ma2019trafficpredict}. In this paper, we employ the attention based cross-agent module to capture explicit agent-to-agent interactions. Even with the increasing number of agents around the ego-vehicle, our cross-agent module can successfully model the interactions between agents, as supported in one of our experiments.

\noindent \textbf{Diverse trajectory forecasting} involves stochastic modeling of future trajectories and sampling diverse predictions based on the model distribution. The Dynamic Bayesian network (DBN) is a common approach without deep generative models, utilized for modelling vehicle trajectories \cite{schulz2018interaction, gindele2015learning, xie2017vehicle} and pedestrian actions \cite{ballan2016knowledge, kooij2014context}. Although the DBN enables the models to consider physical process that generates agent trajectories, the performance is often limited for real traffic scenarios. Most state-of-the-art models utilize deep generative models such as GAN \cite{gupta2018social, sadeghian2019sophie, zhao2019multi} and VAE \cite{lee2017desire} to encourage diverse predictions. However, these approaches mostly focus on generating multiple candidates while less focusing on analyzing the diversity across distributional modes. Recently, sophisticated sampling functions are proposed to tackle this issue, such as Diversity Sampling Function \cite{yuan2019diverse} and Latent Semantic Sampling \cite{huang2020diversitygan}. Despite some promising empirical results, it remains difficult to evaluate both the diversity and admissibility of predictions. In this paper, we tackle the task of diverse trajectory forecasting with a special emphasis on admissibility in dynamic scenes and propose a new task metric that specifically assess models on the basis of these attributes.

\section{Problem Formulation}
\label{sec:problem_definition}

We define the terminology that constitutes our problem. An \textit{agent} is a dynamic on-road object that is represented as a sequence of 2D coordinates, i.e., a spatial position over time. We denote the position for agent $a$ at time $t$ as $S^a_t \in \mathbb{R}^2$, sequence of such positions from $t_1$ to $t_2$ as $S^a_{t_1:t_2}$, and the full sequence as (bold) $\bm{S}^a$. We set $t=0$ as \textit{present}, $t \leq 0$ as \textit{past}, and $t > 0$ as \textit{prediction} or simply,  \textit{pred}. We often split the sequence into two parts, with respect to the \textit{past} and \textit{pred} sub-sequences: we denote these as $\bm{S}^a_{\text{past}}$ and $\bm{S}^a_{\text{pred}}$. In order to clearly distinguish the predicted values from these variables, we use `hats' such as $\hat{S}^a_{t}$ and $\hat{\bm{S}}_{\text{pred}}$. A \textit{scene} is a high-dimensional structured data that describes the present environmental context around the agent. For this, we utilize a bird's eye view array, denoted as $\bm{\Phi} \in \mathbb{R}^{H \times W \times C}$, where $H$ and $W$ are the sizes of field around the agent and $C$ is the channel size of the scene, where each channel consists of distinct information such as the drivable area, position, and distance encodings.

Combining the \textit{scene} and all \textit{agent} trajectories yields an \textit{episode}. In the combined setting, there are a variable number of agents which we denote using bold $\bm{S} \equiv \{\bm{S}^{1}, \bm{S}^{2}, ..., \bm{S}^{A}\}$ and as similarly to other variables, we may split it into two subsets, $\bm{S}_{\text{past}}$ and $\bm{S}_{\text{pred}}$ to represent the past and prediction segments. Since $\bm{S}_{\text{past}}$ and $\bm{\Phi}$ serve as the observed information cue used for the prediction, they are often called \textit{observation} simply being denoted as $\mathcal{O} \equiv \{\bm{S}_{\text{past}}, \bm{\Phi}\}$.

We define \textit{diversity} to be the level of coverage in a model's predictions, across modes in a distribution representing all possible future trajectories. We denote the model distribution as $q(\bm{S}^{a}_{\text{pred}} | \mathcal{O})$ and want the model to generate multiple samples interpreting each sample as an independent \textit{hypothesis} that might have happened, given the same observation.

We also acknowledge that encouraging a model's predictions to be diverse, alone, is not sufficient for accurate and safe output; the model predictions should lie in the support of the true future trajectory distribution $p(\bm{S}_{\text{pred}} | \mathcal{O})$, i.e., predictions should be $admissible$. Given the observation $\mathcal{O}$, it is futile to predict samples around regions that are physically and statistically implausible to reach.

To summarize, this paper addresses the task of \textit{diverse and admissible multi-agent trajectory forecasting}, by modeling multiple modes in the posterior distribution over the \textit{prediction} trajectories, given the observation.

\section{Proposed Approach}
\label{sec:proposed_approach}

We hypothesize that future trajectories of human drivers should follow distributions of multiple modes, conditioned on the scene context and social behaviors of agents. Therefore, we design our model to explicitly capture both agent-to-scene interactions and agent-to-agent interactions with respect to each agent of interest. Through our objective function, we explicitly encourage the model to learn a distribution with multiple modes by taking into account past trajectories and attended scene context.

\subsection{Model Architecture}
\label{sec:model_arch}

\begin{figure}[t]
\centering
\includegraphics[width=10cm]{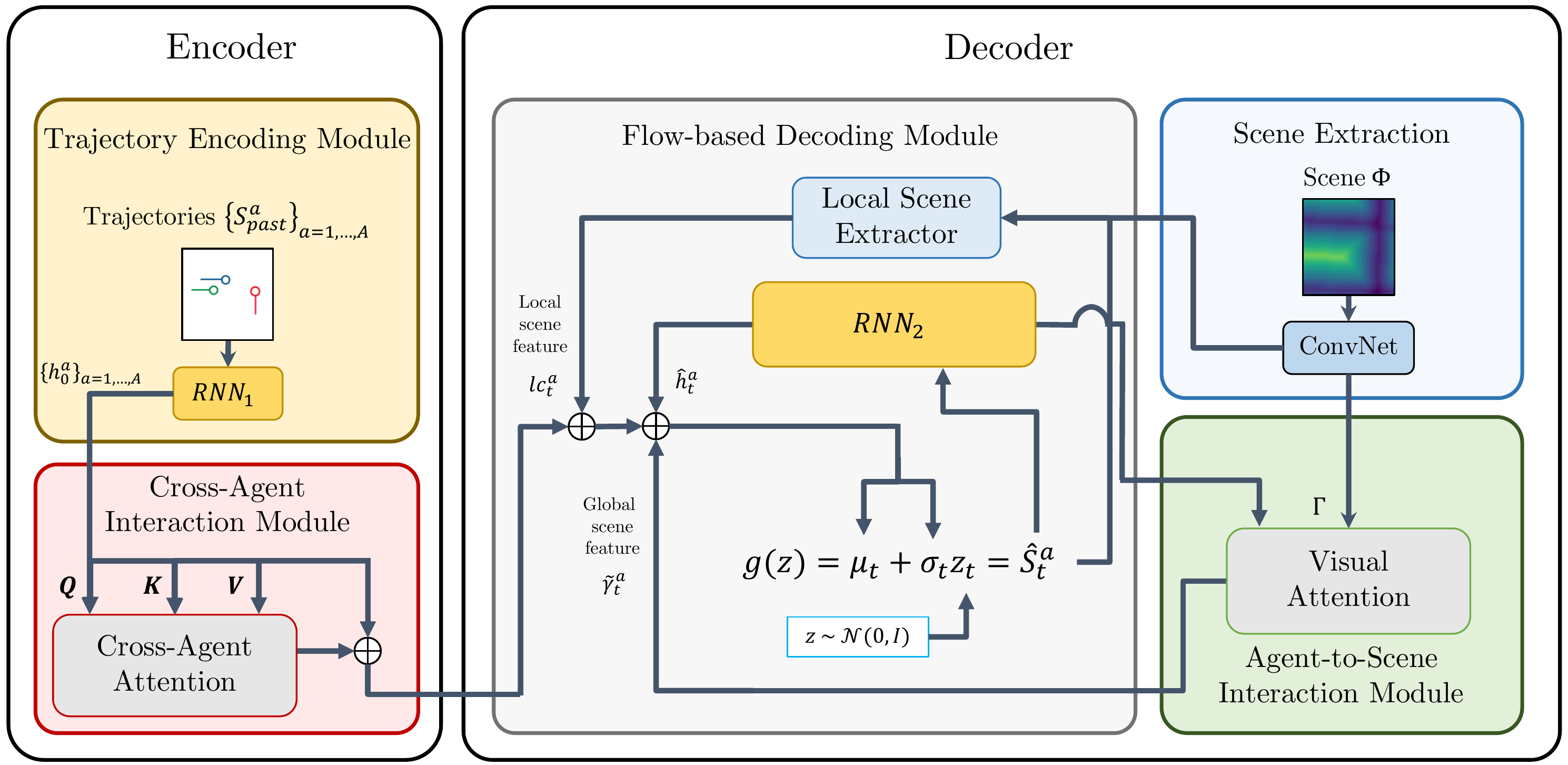}
\caption{Model Architecture. The model consists of an encoder-decoder architecture: the encoder takes as past agent trajectories and calculates cross-agent attention, and the flow-based decoder predicts future trajectories by attending scene contexts for each decoding step.}
\label{fig:model_arch}
\end{figure}

\begin{figure}[!t]
\centering
\subfigure[]{ 
\includegraphics[width=.45\linewidth]{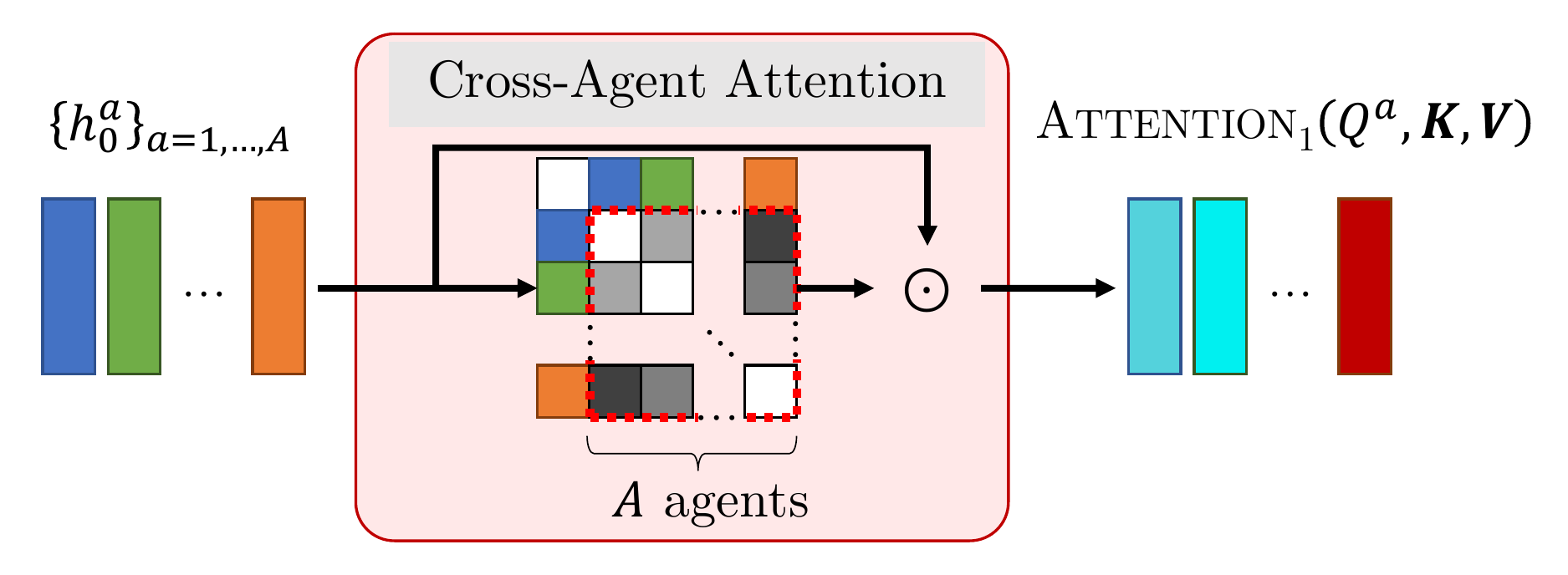}
\label{fig:CrossAgentModule}
}
\subfigure[]{
\includegraphics[width=.45\linewidth]{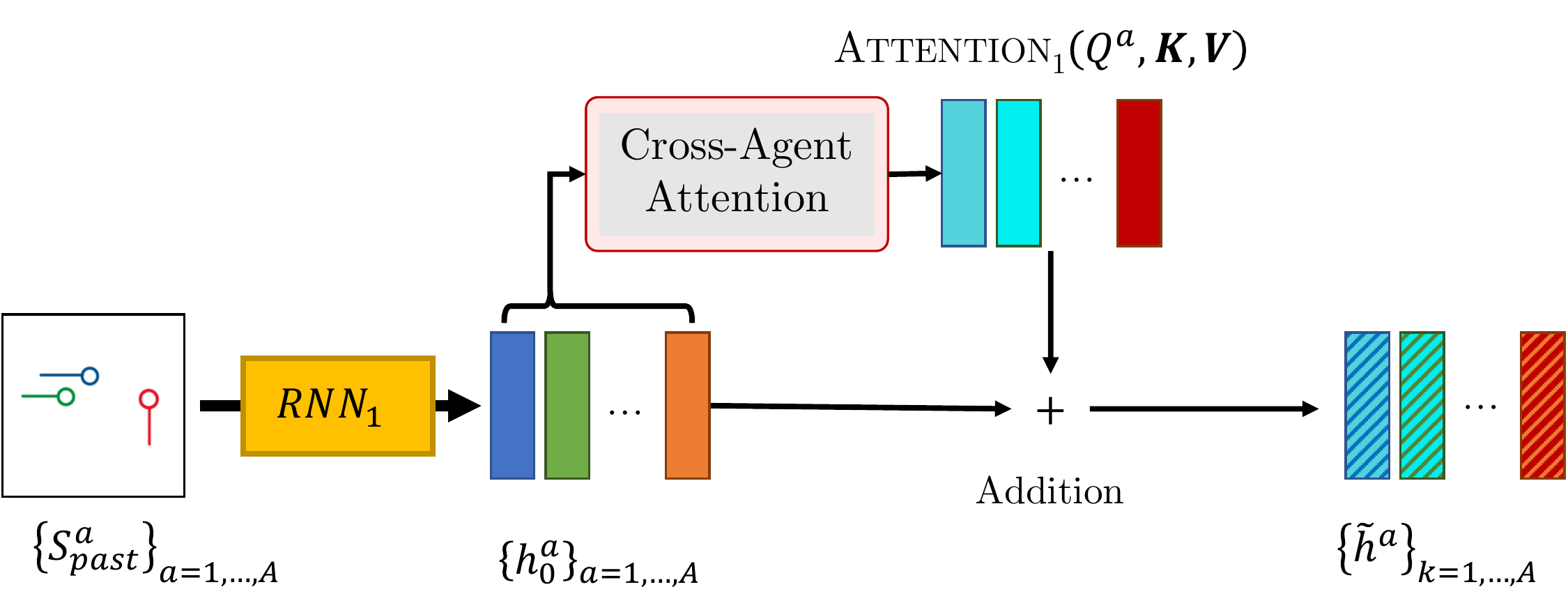}
\label{fig:AgentInteractionModel}
}
\subfigure[]{
\includegraphics[width=.45\linewidth]{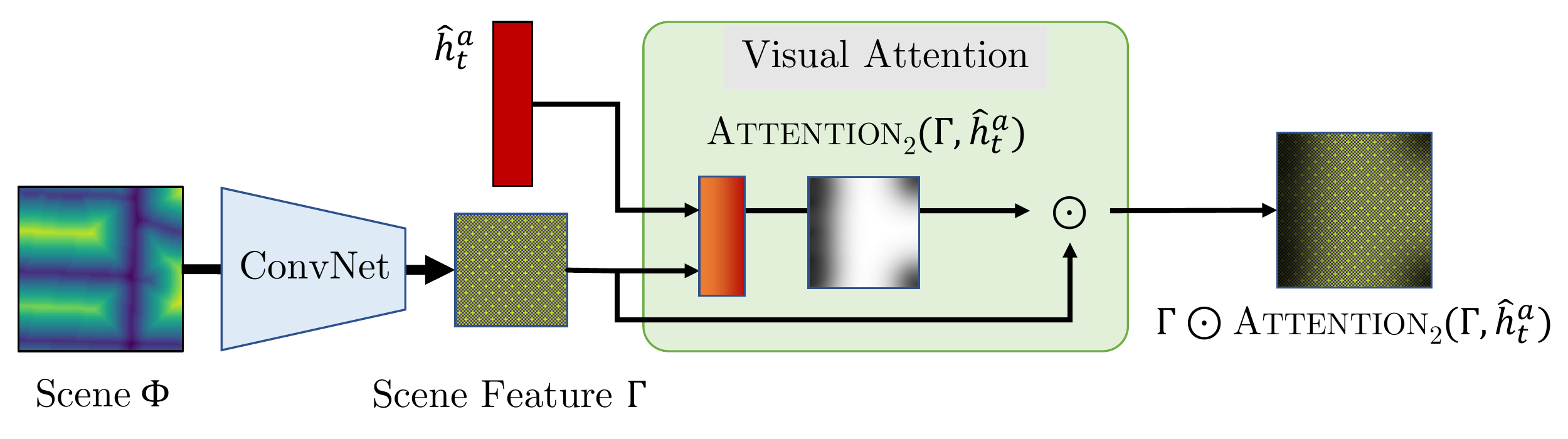}
\label{fig:VisualAttentionModule}
}
\subfigure[]{
\includegraphics[width=.45\linewidth]{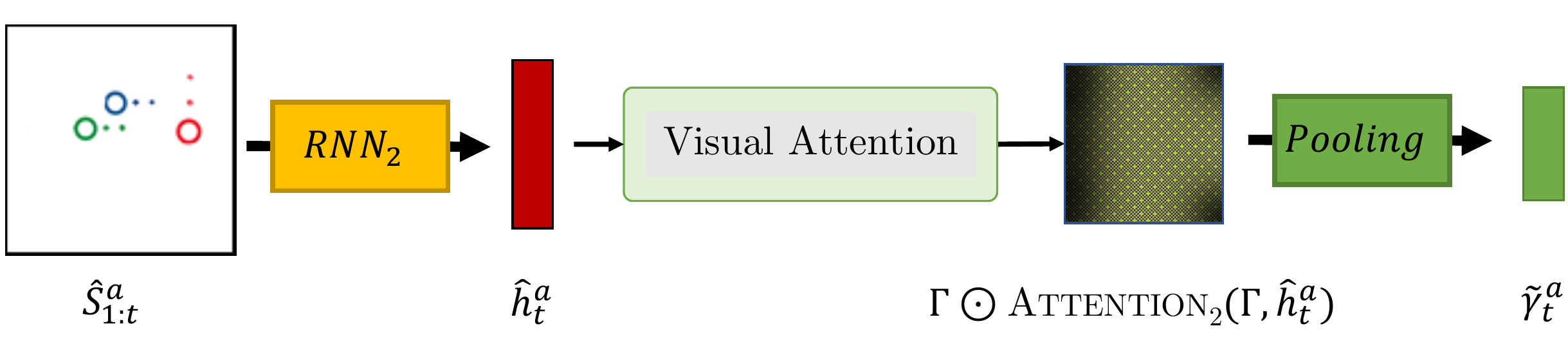}
\label{fig:SceneInteractionModel}
}
\caption{
(a) Cross-agent attention. Interaction between each agent is modeled using attention,
(b) Cross-agent interaction module. Agent trajectory encodings are corrected via cross-agent attention.
(c) Visual attention. Agent-specific scene features are calculated using attention.
(d) Agent-to-scene interaction module. Pooled vectors are retrieved from pooling layer after visual attention.
}
\label{fig:model_modules}
\end{figure}

As illustrated in Fig.~\ref{fig:model_arch}, our model consists of an encoder-decoder architecture. The encoder includes the cross-agent interaction module. The decoder, on the other hand, comprises the agent-to-scene interaction module to capture the scene interactions. Please refer to Fig.~\ref{fig:model_modules} for a detailed illustration of our main proposed modules.

The \textit{encoder} extracts past trajectory encoding for each agent, then calculates and fuses the interaction features among the agents. Given an observation,
we encode each agent's past trajectory 
$S^{a}_{\textrm{past}}$ by feeding it to the trajectory encoding module. The module has the LSTM-based layer $\textsc{RNN}_1$ to summarize the past trajectory. It iterates through the past trajectory with Eq.~(\ref{eqn:rnn1}) and its final output $h_{0}^{a}$ (at \textit{present} $t=0$) is utilized as the agent embedding.
The collection of the embeddings for all agents
is then passed to the cross-agent interaction module, depicted in Fig.~\ref{fig:CrossAgentModule}, which uses \textit{self-attention}~\cite{vaswani2017attention} to generate a cross-agent representation. 
We linearly transform each agent embedding to get a query-key-value triple, $({Q}^{a}, {K}^{a}, {V}^{a})$. Next, we calculate the interaction features through the self-attention layer $\textsc{Attention}_1$.
Finally, the fused agent encoding
$\tilde{\bm{h}} \equiv \{\tilde{h}^{1}, \tilde{h}^{2}, ..., \tilde{h}^{A}\}$
is calculated by adding each attended features to the corresponding embedding (see Eq.~(\ref{eqn:attention1}) and Fig.~\ref{fig:AgentInteractionModel}). 

\begin{gather}
h^{a}_{t} = \textsc{RNN}_{1} (S^a_{t-1}, h^{a}_{t-1})  \label{eqn:rnn1} \\
\tilde{h}^{a} = h_{0}^{a}  + \textsc{Attention}_1 (Q^{a},\bm{K},\bm{V}) \label{eqn:attention1}
\end{gather}

The \textit{decoder} takes the final encoding $\tilde{{h}}^{a}$ and the scene context $\bm{\Phi}$ as inputs. We first extract the scene feature through convolutional neural networks i.e., $\bm{\Gamma} = \textsc{CNN} (\bm{\Phi})$.
The decoder then autoregressively generates the future position $\hat{S}^{a}_{t}$, while referring to both local and global scene context from the agent-to-scene interaction module.
The local scene feature is gathered using bilinear interpolation at the local region of $\bm{\Gamma}$, corresponding to the physical position $\hat{S}^{a}_{t-1}$.
Then, the feature is concatenated with $\tilde{h}^{a}$ and processed thorough FC layers to form the local context $\textit{lc}^{a}_{t}$. We call this part the local scene extractor.
The global scene feature is calculated using \textit{visual-attention}~\cite{xu2015show} to generate weighted scene features, as shown in Fig.~\ref{fig:VisualAttentionModule}. To calculate the attention, we first encode previous outputs $\hat{S}_{1:t-1}^{a}$, using a GRU-based $\textsc{RNN}_2$ in Eq.~(\ref{eqn:rnn2}), whose output is then used to calculate the pixel-wise attention $\tilde{\gamma}_{t}^{a}$ at each decoding step, for each agent in Eq.~(\ref{eqn:attention2}).

\begin{gather}
\hat{h}^{a}_{t} = \textsc{RNN}_{2} (\hat{S}^{a}_{1:t-1}, \hat{h}^{a}_{t-1}) \label{eqn:rnn2} \\
\tilde{\gamma}^{a}_{t} = \textsc{Pool} ( \bm{\Gamma} \odot \textsc{Attention}_{2} (\hat{h}^a_t, \bm{\Gamma}) ) \label{eqn:attention2}
\end{gather}

The flow-based decoding module generates the future position $\hat{S}^{a}_{t}$. The module utilizes \textit{Normalizing Flow}~\cite{rezende2015variational}, a generative modeling method using bijective and differentiable functions; in particular, we choose an autoregressive design~\cite{kingma2016improved, rhinehart2018r2p2, rhinehart2019precog}. We concatenate $\tilde{\gamma}_{t}^{a}$, $\hat{h}^a_t$, and $\textit{lc}^{a}_{t}$ then project them down to 6-dimensional vector using FC layers. We split this vector to $\hat{\mu}_{t} \in \mathbb{R}^2$ and $\hat{\sigma}_{t} \in \mathbb{R}^{2 \times 2}$. Next, we transform a standard Gaussian sample $z_{t} \thicksim \mathcal{N}(\bm{0}, I) \in \mathbb{R}^2$, by the mapping $g_\theta(z_t; \mu_t, \sigma_t) = \sigma_t \cdot z_t + \mu_t = \hat{S}^{a}_{t}$, where $\theta$ is the set of model parameters. `hats' over $\hat{\mu}_t$ and $\hat{\sigma}_t$ are removed, in order to note that they went through the following details. To ensure the positive definiteness, we apply matrix exponential $\sigma_{t} = \textit{expm} ( \hat{\sigma}_{t})$ using the formula in~\cite{bernstein1993some}. Also, to improve the the physical \textit{admissibility} of the prediction, we apply the constraint $\mu_{t} = \hat{S}^{a}_{t-1} + \alpha (\hat{S}^{a}_{t-1} - \hat{S}^{a}_{t-2}) + \hat{\mu}_{t}$, where $\alpha$ is a model degradation coefficient. When $\alpha = 1$, the constraint is equivalent to \textit{Verlet integration}~\cite{verlet1967computer}, used in some previous works~\cite{rhinehart2018r2p2, rhinehart2019precog}, which gives the a perfect constant velocity (CV) prior to the model. However, we found empirically that, the model easily overfits to datasets when the the perfect CV prior is used, and perturbing the model with $\alpha$ prevents overfitting; we choose $\alpha = 0.5$.

Iterating through time, we get the predictive trajectory $\hat{S}^{a}_{\textrm{pred}}$ for each agent. By sampling multiple instances of $z_{\textrm{pred}}$ and mapping them to trajectories, we get various hypotheses of future. Further details of the our network architecture and experiments for the degradation coefficient $\alpha$ are given in the supplementary.

\subsection{Learning}
\label{sec:method:learning}

Our model learns to predict the distribution over the future trajectories of agents present in a given episode. In detail, we focus on predicting the conditional distribution $p(\bm{S}_{\text{pred}} | \mathcal{O})$ where the future trajectory $\bm{S}_{\text{pred}}$ depends on the observation.
As described in the previous sections, our model incorporates a bijective and differentiable mapping 
between standard Gaussian prior $q_0$ and the future trajectory distribution $q_\theta$.
Such technique, commonly aliased `normalizing flow', enables our model not only to generate multiple candidate samples of future, but also to evaluate the ground-truth trajectory with respect to the predicted distribution $q_\theta$ by using the inverse and the change-of-variable formula in Eq.~(\ref{eqn:change-of-variable}).
\begin{equation}
    \label{eqn:change-of-variable}
    q_\theta(S^{a}_{\text{pred}}|\mathcal{O}) = q_0 (g^{-1}(S^{a}_{\text{pred}})) \left| \text{det}(\partial S^{a}_{\text{pred}} / \partial (g^{-1} (S^{a}_{\text{pred}}))   \right|^{-1}
\end{equation}
As a result, our model can simply
close the discrepancy between the predictive distribution $q_\theta$ and the real world distribution $p$ by optimizing our model parameter $\theta$. In particular, we choose to minimize the combination of forward and reverse cross-entropies, also known as `symmetric cross-entropy', between the two distributions in Eq. ~(\ref{eqn:symmetric_ce}). Minimizing the symmetric cross-entropy allows model to learn generating diverse and plausible trajectory, which is mainly used in~\cite{rhinehart2018r2p2}.
\begin{equation}
    \label{eqn:symmetric_ce}
    \min_\theta H(p, q_\theta) + \beta H(q_\theta, p)
\end{equation}
To realize this, we gather the ground-truth trajectories $\bm{S}$ and scene context $\bm{\Phi}$ from the dataset that we assume to well reflect the real distribution $p$, then optimize the model parameter $\theta$ such that 1) the density of the ground-truth future trajectories on top of the predicted distribution $q_\theta$ is maximized and 2) the density of the predicted samples on top of the real distribution $p$ is also maximized as described in Eq.~(\ref{eqn:objective}). Since this objective is fully differentiable with respect to the model parameter $\theta$, we train our model using the stochastic gradient descent.
\begin{equation}
    \label{eqn:objective}
    \min_\theta \E_{\bm{S}_{\text{pred}}, \mathcal{O} \thicksim p} \bigg[ \E_{S^{a}_{t} \in \bm{S}_{\text{pred}}} -\log{q_\theta(S^{a}_{t}|\mathcal{O})} + \beta \E_{\hat{S}^{a}_{t} \thicksim q_\theta} -\log{p(\hat{S}^{a}_{t}|\mathcal{O})} \bigg]
\end{equation}
Such symmetric combination of the two cross-entropies guides our model to predict $q_\theta$ that covers all plausible modes in the future trajectory while penalizing the bad samples that are less likely under the real distribution $p$. However, one major problem inherent in this setting is that we cannot actually evaluate $p$ in practice. In this paper, we propose a new method which approximates $p$ using the drivable-area map and we discuss it in the following subsection~\ref{sec:p_tilled}. Other optimization details are included in the supplementary.

\subsection{The Drivable-area Map and Approximating $p$}
\label{sec:p_tilled}

\begin{figure}[t]
\centering
\includegraphics[width=12cm]{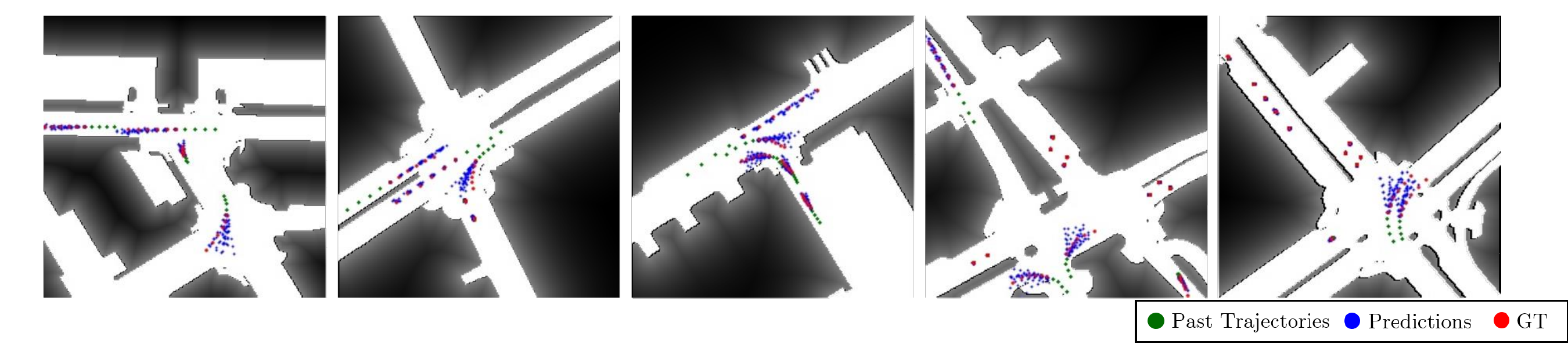}
\caption{
Our map loss and corresponding model predictions. Each pixel on our map loss denotes probability of future trajectories; higher probability values are represented by brighter pixels. Our approach generates diverse and admissible future trajectories. More visualizations of qualitative results are provided in the supplementary.}
\label{fig:model_results}
\end{figure}

We generate a binary mask that denote the drivable-area around the agents. We refer to this feature as the \textit{drivable-area map} and utilize it for three different purposes: 1) deriving the approximated true trajectory distribution $\tilde{p}$, 2) calculating the diversity and admissibility metrics, and 3) building the scene input $\bm{\Phi}$.
Particularly, $\tilde{p}$ is a key component in our training objective, Eq.~(\ref{eqn:symmetric_ce}). Since the reverse cross-entropy penalizes the predicted trajectories with respect to the real distribution, the approximation should not underestimate some region of the real distribution, or diversity in the prediction could be discouraged. Previous works on $\tilde{p}$ utilize the ground-truth (GT) trajectories to model it \cite{rhinehart2018r2p2}. However, there is often only one GT annotation available thus deriving $\tilde{p}$ based on the GT could assign awkwardly low density around certain region. To cope with such problem, our method assumes that every drivable locations are equally probable for future trajectories to appear in and that the non-drivable locations are increasingly less probable, proportional to the distance from the drivable-area. To derive it, we encode the distance on each non-drivable location using the distance transform on the drivable-area maps, then apply softmax over the transformed map to constitute it as a probability distribution. The visualizations of the $\tilde{p}$ are available in Fig.~\ref{fig:model_results}. Further details on deriving $\tilde{p}$ and the scene context input $\bm{\Phi}$, as well as additional visualizations and qualitative results are given in the supplementary.

\section{Experimental Setup}
\label{sec:experiments}

The primary goal in the following experiments is to evaluate our model, baselines, and ablations on the following criteria|(i) Leveraging mechanisms that explicitly model agent-to-agent and agent-to-scene interactions (experiments 1 and 2). (ii) Producing diverse trajectory predictions, while obeying admissibility constraints on the trajectory candidates, given different approximation methods for the true trajectory distribution $p$ (experiment 3). (iii) Remaining robust to an increasing number of agents in the scene (agent complexity; experiment 4). (iv) Generalizing to other domains (experiment 5).

\subsection{Dataset}
\label{sec:dataset}

We utilize two real world datasets to evaluate our model and the baselines: \textsc{nuScenes} tracking \cite{caesar2020nuscenes} and \textsc{Argoverse} motion forecasting \cite{chang2019argoverse}. \textsc{nuScenes} contains 850 different real-world driving scenarios, where each spanning 20 seconds of frames and 3D box annotations for the surrounding objects. It also provides drivable-area maps. Based on this setting, we generate trajectories by associating the box annotations of the same agents. While \textsc{nuScenes} provides trajectories for realistic autonomous driving scenarios, the number of episodes is limited around 25K. On the other hand, \textsc{Argoverse} provides around 320K episodes from the real world driving along with the drivable-area maps. \textsc{Argoverse} presents independent episodes spanning only 5s (2s for the past and 3s for the prediction), rather than providing long, continuing scenarios as in \textsc{nuScenes}. However, this setting suffices to test our method and we evaluate baselines and our models on these two real-world datasets to provide complementary validations of each model’s diversity, admissibility, and generalizing performance across domains.

In order to make the experimental setup of \textsc{nuScenes} similar to \textsc{Argoverse}, we crop each sequence to be 5 seconds long in total; 3 seconds for prediction and 2 seconds for observation, with a sampling rate of 2 Hz. We also employ Kalman smoothing to reduce noise and impute missing points in trajectories. Further details in the data processing are available in the supplementary.

\subsection{Baseline Models}
\label{subsec:baseline_models}

\textbf{Deterministic baselines:} We compare three deterministic models with our approach, to examine our model's ability to capture agent-to-agent interaction:
\textit{LSTM-based encoder-decoder} (\textsc{LSTM}), \textit{LSTM with convolutional social pooling} \cite{deo2018convolutional} (\textsc{CSP}), and a deterministic version of \textit{multi-agent tensor fusion} (\textsc{MATF-D})~\cite{zhao2019multi}. For our deterministic model, we use an LSTM with our cross-agent attention module in the encoder, which we refer to as the \textit{cross-agent attention model} (\textsc{CAM}). Because each model is predicated on an LSTM component, we set the capacity to be the same in all cases, to ensure fair comparison. 

\noindent \textbf{Stochastic baselines:} We experiment three stochastic baselines. Our first stochastic baseline is a model based on a Variational Autoencoder structure, (\textsc{DESIRE}) \cite{lee2017desire}, which utilizes scene contexts and an iterative refinement process. The second baseline model is a Generative Adversarial Network version of \textit{multi-agent tensor fusion} (\textsc{MATF-GAN})~\cite{zhao2019multi}. Our third baseline is the Reparameterized Pushforward Policy (\textsc{R2P2-MA})~\cite{rhinehart2019precog} which is a modified version of R2P2~\cite{rhinehart2018r2p2} for multi-agent prediction. To validate our model's ability to extract scene information and generate diverse trajectories, multiple versions of our models are tested. While these models can be used as stand-alone models to predict diverse trajectories, comparison amongst these new models is equivalent to an ablation study of our final model. \textsc{CAM-NF} is a CAM model with a flow-based decoder. \textsc{Local-CAM-NF} is \textsc{CAM-NF} with local scene features. \textsc{Global-CAM-NF} is \textsc{Local-CAM-NF} with global scene features. Finally, \textsc{AttGlobal-CAM-NF} is \textsc{Global-CAM-NF} with agent-to-scene attention, which is our main proposed model.

\begin{figure}[t]
\centering
\includegraphics[width=10cm]{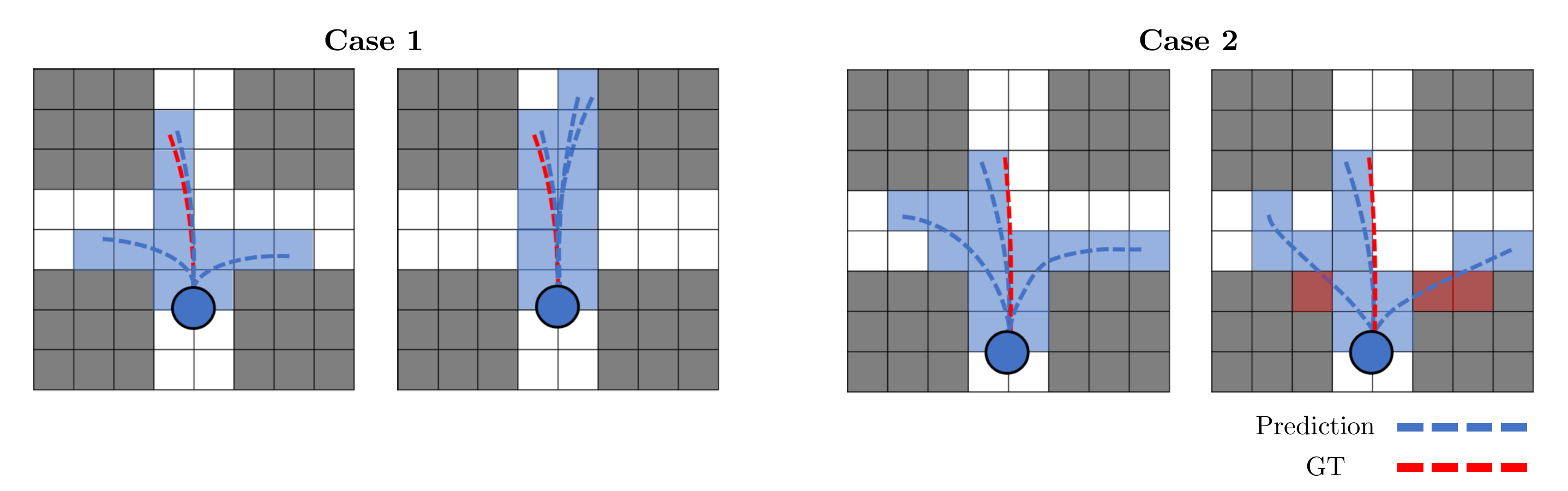}
\caption{We motivate the need for multiple metrics, to assess diversity and admissibility. Case 1: DAO measures are equal, even though predictions have differing regard for the modes in the posterior distribution. Case 2: \textsc{rF} measures are equal, despite differing regard for the cost of leaving the drivable area. In both cases, it is important to distinguish between conditions|we do this by using \textsc{DAO}, \textsc{rF}, and \textsc{DAC} together.}
\label{fig:metric_cases}
\end{figure}

\subsection{Metrics}
\label{sec:metrics}

We define multiple metrics that provide a thorough interpretation about the behavior of each model in terms of precision, diversity, and admissibility. To evaluate precision, we calculate Euclidean errors: $\textsc{ADE}$ (\textit{average displacement error}) and $\textsc{FDE}$ (\textit{final displacement error}), or $\textsc{Error}$ to denote both. To evaluate multiple hypotheses, we use the average and the minimum $\textsc{Error}$ among $K$ hypotheses: $\textsc{avgError} = \frac{1}{k} \sum_{i=1}^{k} \textsc{Error}^{(i)}$ and $\textsc{minError} = \min \{ \textsc{Error}^{(1)}, ..., \textsc{Error}^{(k)} \}$. A large $\textsc{avgError}$ implies that predictions are spread out, and a small $\textsc{minError}$ implies at least one of predictions has high precision. From this observation, we define a new evaluation metric that capture diversity in predictions: the ratio of $\textsc{avgFDE}$ to $\textsc{minFDE}$, namely $\textsc{rF}$.
$\textsc{rF}$ is robust to the variability of magnitude in velocity in predictions hence provides a handy tool that can distinguish between predictions with multiple modes (diversity) and predictions with a single mode (perturbation).

\begin{equation}
    \label{measure:RF}
    \textrm{Ratio of \textsc{avgFDE} to \textsc{minFDE} (\textsc{rF})} = \frac{\textsc{avgFDE}}{\textsc{minFDE}}
\end{equation}

\begin{equation}
    \label{measure:DAO}
    \textrm{Drivable Area Occupancy (\textsc{DAO})} = \frac{\textrm{count}(\textrm{traj}_{\textrm{pix}})}{\textrm{count}(\textrm{driv}_{\textrm{pix}})}
\end{equation}

\begin{figure}[t]
\centering
\includegraphics[width=10cm]{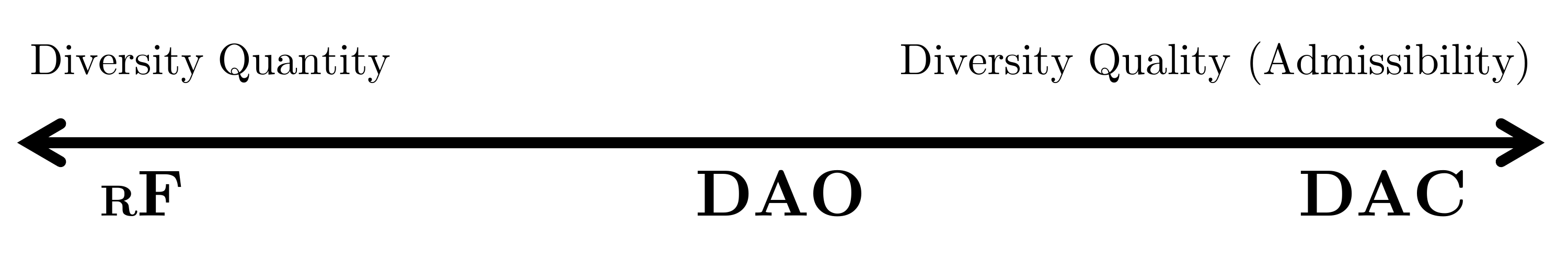}
\caption{Metric quality spectrum. Our newly proposed metrics: \textsc{rF} measures the spread of predictions in Euclidean distance, \textsc{DAO} measures diversity in predictions that are only admissible. \textsc{DAC} measures extreme off-road predictions that defy admissibility.}
\label{fig:metric_diagram}
\end{figure}

We also report performance on additional metrics that are designed to capture diversity and admissibility in predictions.
While \textsc{rF} measures the spread of predictions in Euclidean distance, \textsc{DAO} measures diversity in predictions that are only admissible, and \textsc{DAC} measures extreme off-road predictions that defy admissibility. We follow~\cite{chang2019argoverse} in the use of \textit{Drivable Area Count} (\textsc{DAC}), $\textsc{DAC} = \frac{k-m}{k}$, where $m$ is the number of predictions that go out of the drivable area and $k$ is the number of hypotheses per agent. Next, we propose a new metric, \textit{Drivable Area Occupancy} (\textsc{DAO}), which measures the proportion of pixels that predicted trajectories occupy in the drivable-area. Shown in Eq.~(\ref{measure:DAO}), $\textrm{count}(\textrm{traj}_{\textrm{pix}})$ is the number of pixels occupied by predictions and $\textrm{count}(\textrm{driv}_{\textrm{pix}})$ is the total number of pixels of the drivable area, both within a pre-defined grid around the ego-vehicle.

\textsc{DAO} may seem a reasonable standalone measure of capturing both diversity and admissibility, as it considers diversity in a reasonable region of interest. However, \textsc{DAO} itself cannot distinguish between \textit{diversity} (Section~\ref{sec:problem_definition}) and arbitrary stochasticity in predictions, as illustrated by Case 1 in Fig.~\ref{fig:metric_cases}: although \textsc{DAO} measures of both predictions are equal, the causality behind each prediction is different and we must distinguish the two. \textsc{rF} and \textsc{DAO} work in a complementary way and we, therefore, use both for measuring diversity. To assure the \textit{admissibility} of predictions, we use \textsc{DAC} which explicitly counts off-road predictions, as shown by Case 2 in Fig.~\ref{fig:metric_cases}. As a result, assessing predictions using \textsc{DAO} along with \textsc{rF} and \textsc{DAC} provides a holistic view of the quantity and the quality of diversity in predictions; the characteristics of each metric is summarized in Fig.~\ref{fig:metric_diagram}.

For our experiments, we use \textsc{minADE} and \textsc{minFDE} to measure \textit{precision}, and use \textsc{rF}, \textsc{DAC}, and \textsc{DAO} to measure both \textit{diversity} and \textit{admissibility}. Due to the nature of \textsc{DAO}, where the denominator in our case is the number of overlapping pixels in a $224 \times 224$ grid, we normalize it by multiplying by $10,000$ when reporting results. For the multi-agent experiment (shown in Table~\ref{tab:multiagent}), relative improvement (\textsc{RI}) is calculated as we are interested in the relative improvement as the number of agents increases. For all metrics, the number of trajectory hypotheses should be set equally for fair comparison of models. If not specified, the number of hypotheses $k$ is set to 12 when reporting the performance metrics.


\section{Results and Discussion}
\label{sec:results}

In this section, we show experimental results on numerous settings including the comparison with the baseline, and ablation studies of our model. We first show the effect of our cross-agent interaction module and agent-to-scene interaction module on the model performance, then we analyze the performance with respect to different numbers of agents, and other datasets. All experiments are measured with \textsc{minADE}, \textsc{minFDE}, \textsc{rF}, \textsc{DAC}, and \textsc{DAO} for holistic interpretation.

\noindent \textbf{Effectiveness of cross-agent interaction module:} We show the performance of one of our proposed models $\textsc{CAM}$, which utilizes our cross-agent attention module, along with three deterministic baselines as shown in Table~\ref{tab:cross_agent_exp}. $\textsc{CSP}$ models the interaction through layers of convolutional networks, and the interaction is implicitly calculated within the receptive field of convolutional layers. $\textsc{MATF-D}$ is an extension of convolutional social pooling with scene information. \textsc{CAM} explicitly defines the interaction between each agent by using attention. The result shows that \textsc{CAM} outperforms other baselines in both \textsc{minADE} and \textsc{minFDE}, indicating that the explicit way of modeling agent-to-agent interaction performs better in terms of precision than an implicit way of modeling interaction using convolutional networks used in \textsc{CSP} and \textsc{MATF-D}. Interestingly, \textsc{CAM} outperforms \textsc{MATF-D} that utilizes scene information. 
This infers that the cross-agent interaction module has ability to learn the structure of roads and permissible region given the trajectories of surrounding agents.

\begin{table}[ht]
\caption{Deterministic models on \textsc{nuScenes}. Our proposed model outperforms the existing baselines.}
\begin{center}
\scriptsize
\begin{tabular}{lcccccccccccc}
\toprule
Model & & \textsc{minADE} $(\downarrow)$ & & \textsc{minFDE} $(\downarrow)$ \\
\hline
\shortstack{\textsc{LSTM}} & & \shortstack{1.186} & & \shortstack{2.408} \\
\shortstack{\textsc{CSP}}  & & \shortstack{1.390} & & \shortstack{2.676} \\
\shortstack{\textsc{MATF-D}~\cite{zhao2019multi}}  & & \shortstack{1.261} & & \shortstack{2.538} \\
\hline
\shortstack{\textsc{CAM (ours)}} & & \shortstack{\textbf{1.124}} & & \shortstack{\textbf{2.318}}\\
\bottomrule
\end{tabular}
\end{center}
\label{tab:cross_agent_exp}
\end{table}
\vspace{-15mm}
\begin{table}[ht]
\caption{Stochastic models on \textsc{nuScenes}. Our models outperform the existing baselines, achieving the best precisions, diversity, and admissibility. Improvements indicated by arrows.
}
\begin{center}
\scriptsize
\begin{tabular}{lccccc}
\toprule
Model & \textsc{minADE} $(\downarrow)$ & \textsc{minFDE} $(\downarrow)$ & \textsc{rF} $(\uparrow)$ & \textsc{DAO} $(\uparrow)$ & \textsc{DAC} $(\uparrow)$ \\
\hline
\textsc{DESIRE~\cite{lee2017desire}}  & 1.079 & 1.844 & 1.717 & 16.29 & 0.776\\ \textsc{MATF-GAN}~\cite{zhao2019multi}  & 1.053 & 2.126 & 1.194 & 11.64 & 0.910\\
\textsc{R2P2-MA}~\cite{rhinehart2018r2p2}  & 1.179 & 2.194 & 1.636 & \textbf{25.65} & 0.893\\

\hline

\textsc{CAM-NF (ours)} & 0.756 & 1.381 & 2.123 & 23.15 & 0.914\\
\textsc{Local-CAM-NF (ours)} & 0.774 & 1.408 & 2.063 & 22.58 & \textbf{0.921}\\
\textsc{Global-CAM-NF (ours)} & 0.743 & 1.357 & 2.106 & 22.65 & \textbf{0.921} \\
\textsc{AttGlobal-CAM-NF (ours)} & \textbf{0.639} & \textbf{1.171} & \textbf{2.558} & 22.62 & 0.918 \\

\bottomrule 
\end{tabular}
\end{center}
\label{tab:scene_extraction_exp}
\end{table}
\vspace{-5mm}

\label{exp:agent-to-scene_exp}
\noindent \textbf{Effectiveness of agent-to-scene interaction module:} The performance of stochastic models is compared in Table~\ref{tab:scene_extraction_exp}. We experiment with removing scene processing operations in the decoder to validate the importance of our proposed agent-to-scene interaction module. As mentioned previously, generating multiple modes of sample requires a strong scene processing module and a diversity-oriented decoder. Our models achieves the best precision.
\textsc{MATF-GAN} has a small $\textsc{rF}$ showing that predictions are unimodal, while other models such as VAE-based model (\textsc{DESIRE}) and flow-based models (\textsc{R2P2-MA} and \textsc{ours}) show spread in their predictions. \textsc{R2P2-MA} shows the highest \textsc{DAO}. We note that \textsc{ours} has a comparable \textsc{DAO} while keeping the highest \textsc{rF} and \textsc{DAC}, indicating that our models exhibit diverse and admissible predictions by accurately utilizing scene context.

\label{exp:p_loss}
\noindent \textbf{Effectiveness of new} $\tilde{\bm{p}}$: We compare two different versions of our \textsc{AttGlobal-CAM-NF}. One is trained using mean squared error (MSE) between $\bm{S}_{pred}$ and $\hat{\bm{S}}_{pred}$ as an example of annotation-based approximation for $p$, while the other is trained with our drivable area-based (annotation-free) approximation of $p$
in Table~\ref{tab:p_loss_exp}. Using our new approximation in training shows superior results in most of the reported metrics. In particular, the precision and the diversity of predictions increases drastically as reflected in \textsc{minError}, \textsc{DAO}, and \textsc{rF} while \textsc{DAC} remains comparable. 
Thus our $\tilde{p}$ considers admissibility while improving precision and diversity i.e drivable-area related approximate enhances the estimate on additional trajectories over the most probable one.

\begin{table}[t]
\caption{Training \textsc{AttGlobal-CAM-NF} using the proposed annotation-free $\tilde{p}$ outperforms the annotation-dependent counterpart (MSE) in \textsc{nuScenes}.}
\centering
\scriptsize
\begin{tabular}{lcccccccccccc}
\toprule
Model & & \textsc{minADE} $(\downarrow)$ & & \textsc{minFDE} $(\downarrow)$ & & \textsc{rF} $(\uparrow)$ & & \textsc{DAO} $(\uparrow)$ & & \textsc{DAC} $(\uparrow)$ \\
\hline
\shortstack{\textsc{AttGlobal-CAM-NF} (MSE)} & & \shortstack{0.735} & & \shortstack{1.379} & & \shortstack{1.918} & & \shortstack{21.48} & & \shortstack{\textbf{0.924}} \\
\shortstack{\textsc{AttGlobal-CAM-NF}} & & \shortstack{\textbf{0.638}} & & \shortstack{\textbf{1.171}} & & \shortstack{\textbf{2.558}} & & \shortstack{\textbf{22.62}} & & \shortstack{0.918} \\
\bottomrule 
\end{tabular}
\label{tab:p_loss_exp}
\end{table}

\begin{table}[t]
\caption{Multi-agent experiments on \textsc{nuScenes} (\textsc{minFDE}). RI denotes relative improvements of \textsc{minFDE} between 10 and 1 agent. Our approach best models multi-agent interactions.
} 
\centering
\scriptsize
\begin{tabular}{lcccccccccccc}
\toprule
Model & & 1 agent & & 3 agents & & 5 agents & & 10 agents & & \textsc{RI(1-10)}\\ 
\hline
\shortstack{\textsc{LSTM}} & & \shortstack{2.736} & & \shortstack{2.477}  & & \shortstack{2.442} & & \shortstack{2.268} & & \shortstack[r]{17.1\%} \\
\shortstack{\textsc{CSP}}  & & \shortstack{2.871}  & & \shortstack{2.679}   & & \shortstack{2.671} & & \shortstack{2.569} & & \shortstack[r]{10.5\%}\\
\shortstack{\textsc{DESIRE}~\cite{lee2017desire}}  & & \shortstack{2.150}  & & \shortstack{1.846}   & & \shortstack{1.878} & & \shortstack{1.784} & & \shortstack[r]{17.0\%}\\
\shortstack{\textsc{MATF GAN}~\cite{zhao2019multi}}  & & \shortstack{2.377}  & & \shortstack{2.168}   & & \shortstack{2.150} & & \shortstack{2.011} & & \shortstack[r]{15.4\%} \\
\shortstack{\textsc{R2P2-MA}~\cite{rhinehart2018r2p2}}  & & \shortstack{2.227}  & & \shortstack{2.135}   & & \shortstack{2.142} & & \shortstack{2.048} & & \shortstack[r]{8.0\%} \\

\hline
\shortstack{\textsc{AttGlobal-CAM-NF (ours)}} & & \shortstack{\textbf{1.278}}  & & \shortstack{\textbf{1.158}} & & \shortstack{\textbf{1.100}} & & \shortstack{\textbf{0.964}} & & \shortstack{\textbf{24.6\%}} \\
\bottomrule
\end{tabular}
\label{tab:multiagent}
\end{table}

\begin{table}[t]
\caption{Results of baseline models (upper partition) and our proposed models (lower partition). \textsc{AttGlobal-CAM-NF} is our full proposed model and others in the lower partition are the ablations. The metrics are abbreviated as follows: \textsc{minADE}(\textbf{A}), \textsc{minFDE}(\textbf{B}),  \textsc{rF}(\textbf{C}), \textsc{DAO}(\textbf{D}), \textsc{DAC}(\textbf{E}). Improvements indicated by arrows.}

\setlength{\tabcolsep}{0.1pt}
\scriptsize
\centering
\begin{tabular}{p{2.9cm}p{0.9cm}p{0.9cm}p{0.9cm}p{0.9cm}p{0.25cm}p{0.9cm}p{0.9cm}p{0.9cm}p{0.9cm}p{0.9cm}p{0.9cm}}

    \toprule

    \multirow{ 3}{*}{Model} & \multicolumn{5}{c}{\textsc{Argoverse}} &&  \multicolumn{5}{c}{\textsc{nuScenes}} \\

    \cmidrule(r{12pt}){2-7} \cmidrule(r{4pt}){8-12} & A~($\downarrow$) & B~($\downarrow$) & C~($\uparrow$)* & D~($\uparrow$)* & E~($\uparrow$)* && A~($\downarrow$) & B~($\downarrow$) & C~($\uparrow$)* & D~($\uparrow$)* & E~($\uparrow$)* \\
    
    \midrule

    {\textsc{LSTM}} & 1.441 & 2.780 & 1.000 & 3.435 & 0.959 && 1.186 & 2.408 &  1.000 &  3.187 & 0.912 \\

    {\textsc{CSP}} & 1.385 & 2.567 & 1.000 & 3.453 & 0.963 && 1.390 & 2.676 &  1.000 &  3.228 & 0.900 \\

    {\textsc{MATF-D}~\cite{zhao2019multi}} & 1.344 & 2.484 & 1.000 & 3.372 & 0.965 && 1.261 & 2.538 & 1.000 & 3.191 & 0.906 \\

    {\textsc{DESIRE}~\cite{lee2017desire}} & 0.896 & 1.453 & 3.188 & 15.17 & 0.457 && 1.079 & 1.844 & 1.717 & 16.29 & 0.776 \\
 
    {\textsc{MATF-GAN}~\cite{zhao2019multi}} & 1.261 & 2.313 & 1.175 &  11.47 & 0.960 && 1.053 & 2.126 & 1.194 & 11.64 & 0.910 \\

    {\textsc{R2P2-MA}~\cite{rhinehart2019precog}} & 1.108 & 1.771 & 3.054 & \textbf{37.18} & 0.955 && 1.179 & 2.194 & 1.636 & \textbf{25.65} & 0.893 \\

    \midrule
    
    {\textsc{CAM}}& 1.131 & 2.504 & 1.000 & 3.244 & \textbf{0.973} && 1.124 & 2.318 & 1.000 & 3.121 & \textbf{0.924} \\
    
    {\textsc{CAM-NF}} & 0.851 & 1.349 & 2.915 & 32.89 & 0.951 && 0.756 & 1.381 & 2.123 & 23.15 & 0.914 \\
    
    {\textsc{Local-CAM-NF}} & 0.808 & 1.253 & 3.025 & 31.80 & 0.965 &&  0.774 & 1.408 & 2.063 & 22.58 & 0.921 \\
    
    {\textsc{Global-CAM-NF}} & 0.806 & 1.252 & 3.040 & 31.59 & 0.965 && 0.743 & 1.357 & 2.106 & 22.65 & 0.921 \\
    
    {\textsc{AttGlobal-CAM-NF}} & \textbf{0.730} & \textbf{1.124} &  \textbf{3.282} & 28.64 & 0.968 && \textbf{0.639} & \textbf{1.171} &  \textbf{2.558} & 22.62 & 0.918 \\
    
    \midrule
    
\end{tabular}
\label{tab:main_experiment}
\end{table}

\label{exp:complexity_multiagent}
\noindent \textbf{Complexity from number of agents:} We experiment with varying number of surrounding agents as shown in Table~\ref{tab:multiagent}. Throughout all models, the performance increases as the number of agents increases even though we observe that many agents in the surrounding do not move significantly. In terms of relative improvement \textsc{RI}, as calculated between 1 agent and 10 agents, our model has the most improvement, indicating that our model makes the most use of the fine-grained trajectories of surrounding agents to generate future trajectories.

\label{exp:generalizability_dataset}
\noindent \textbf{Generalizability across datasets:} We compare our model with baselines extensively across another real world dataset \textsc{Argoverse} to test generalization to different environments. We show results in Table~\ref{tab:main_experiment} where we outperform or achieve comparable results as compared to the baselines. For \textsc{Argoverse}, we additionally outperform MFP3~\cite{tang2019multiple} in \textsc{minFDE} with 6 hypotheses: our full model shows a \textsc{minFDE} of 0.915, while MFP3 achieves 1.399.

\section{Conclusion}
\label{sec:conclusion}

In this paper, we tackled the problem of generating diverse and admissible trajectory predictions by understanding each agent's multimodal context. We proposed a model that learns agent-to-agent interactions and agent-to-scene interactions using attention mechanisms, resulting in better prediction in terms of precision, diversity, and admissibility. We also developed a new approximation method that provides richer information about the true trajectory distribution and allows more accurate training of flow-based generative models. Finally, we present new metrics that provide a holistic view of the quantity and the quality of diversity in prediction, and a \textsc{nuScenes} trajectory extraction code to support future research in diverse and admissible trajectory forecasting.

\bibliographystyle{splncs04}
\bibliography{references}

\newpage
\input{supplementary}

\end{document}

%% file: supplementary.tex
\appendix

\section{Additional Experimental Details}
\label{app:sec:experiments}

\begin{figure}[t]
\centering
\includegraphics[width=0.9\linewidth]{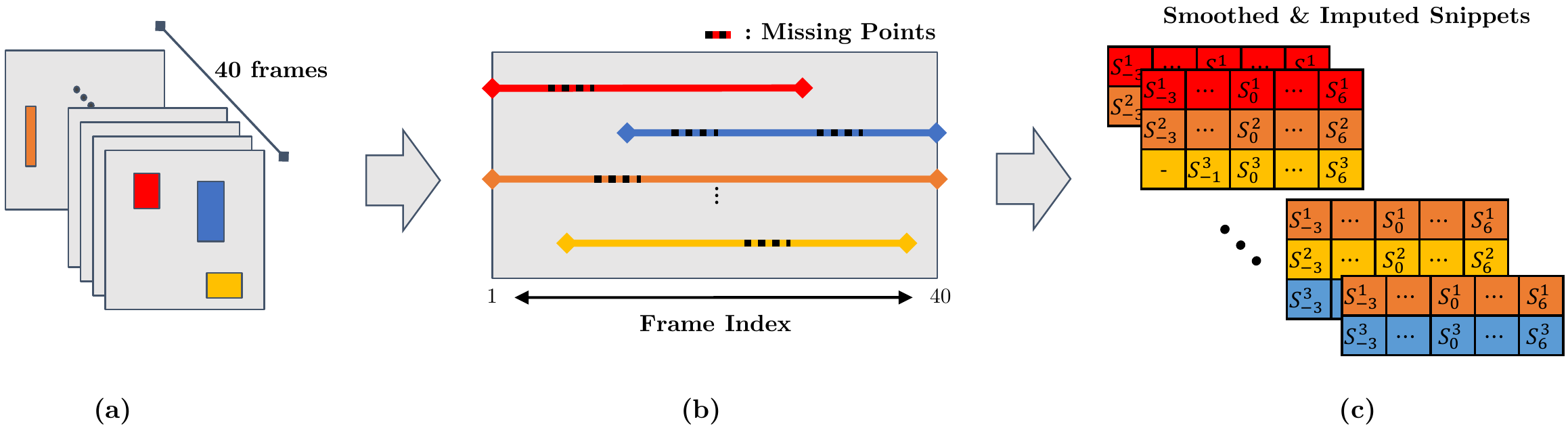}
\caption{Trajectory association, filtering and imputation process. (a), (b), (c) depict the bounding boxes in different frames, the associated and concatenated sequences, and the final trajectory snippets, respectively. Each agent is assigned a different color.}
\label{fig:data_extraction}
\end{figure}

\subsection{nuScenes Trajectory Extraction Process}
\label{app:sec:nuscenes}

nuScenes tracking dataset contains 850 different real-world driving records, each of which spans 40 frames (20 seconds) of LiDAR point-cloud data, RGB camera images, ego-vehicle pose records, and 3D bounding-box annotations of the surrounding vehicles, pedestrians, animals, etc. It also provides a map API that gives an access to the drivable area information. Based on this setting, we generate trajectories by associating the bounding boxes of the same agents, using the agent IDs available in the dataset. The resultant sequences, however, include noise and missing points, due to annotation errors and occlusion issues, as depicted in Fig.~\ref{fig:data_extraction}(b). To combat this, we employ \textit{Kalman smoothing} \cite{yu2004deriv} with the constant velocity linear model in Eq.~(\ref{eqn:cv_model}), in order to filter the noise and impute missing points. When initializing the Kalman filter, we utilize \textit{Expectation-Maximization} (EM) \cite{abbeel2011max} to fit the initial states and the covariance parameters (transition and emission), with respect to the sequences.

\begin{gather}
    \begin{bmatrix} x_{t+1}\\
    y_{t+1} \\
    z_{t+1} \\
    \dot{x}_{t+1}\\
    \dot{y}_{t+1}\\
    \dot{z}_{t+1} \end{bmatrix}
    = 
    \begin{bmatrix}
        1~~ & 0~~ & 0~~ & 0.5 & 0~ & 0 \\
        0~~ & 1~~ & 0~~ & 0~ & 0.5 & 0 \\
        0~~ & 0~~ & 1~~ & 0~ & 0~ & 0.5 \\
        0~~ & 0~~ & 0~~ & 1~ & 0~ & 0 \\
        0~~ & 0~~ & 0~~ & 0~ & 1~ & 0 \\
        0~~ & 0~~ & 0~~ & 0~ & 0~ & 1 \\
    \end{bmatrix} \begin{bmatrix} x_{t}\\
    y_{t}\\
    z_{t} \\
    \dot{x}_{t}\\
    \dot{y}_{t} \\
    \dot{z}_{t} \end{bmatrix} \label{eqn:cv_model} 
\end{gather}

\noindent Next, we construct the episodes for our trajectory dataset. The lengths of the past and prediction segments are set to 2 and 3 seconds (4 and 6 frames), respectively. As a result, an individual episode should contain a snippet of length 5 seconds (10 frames) from the smoothed trajectory, with a maximum of 30 samples extracted from one episode. For each snippet, we normalize the coordinate system, such that the ego-pose at the present time (4th frame) is placed at the origin $(0, 0)$.

For each episode, we generate the road mask of dimension $224 \times 224$ that covers $112 \times 112 (m^2)$ area around the ego-pose, at the present time. The map API enables access to the drivable area information around the spot where the dataset is collected. Based on the ego-pose, we query the drivable area and save the returned information into a binary mask that is assigned $1$'s at the pixels belonging to the drivable area, and $0$'s at the other pixels.

\subsection{Argoverse Motion Forecasting Data}
\label{app:sec:argoverse}
Argoverse motion forecasting contains about 320,000 episodes, each of which spans 50 frames (5 seconds), along with a map API that gives an access to the drivable area information. All of the episodes are pre-processed for the trajectory forecasting task and, hence, additional extraction processes (such as association, smoothing, and imputation) are unnecessary. However, to condition the data samples to be as similar to the samples in the nuScenes trajectory (Section \ref{app:sec:nuscenes}) as possible, we down-sample each episode to episode lengths of 10 frames (5 seconds). We also normalize the coordinate system, such that the ego-agent at the present time is placed at the origin. As with nuScenes, we generate for each episode the road mask of dimension $224 \times 224$ which covers a $112 \times 112 (m^2)$ area around the ego-agent at the present time.

\subsection{CARLA Trajectory Extraction Process and Experiments}
\label{app:sec:carla}

\noindent CARLA \cite{dosovitskiy1711carla} is a vehicle and pedestrian behavior simulator wherein the agents follow the physical laws of motions driven by Unreal Engine. In order to validate the generalizability of our models, we synthesize trajectory forecasting data using CARLA vehicles and record the trajectory for all simulation time-steps. We make sure the data format is consistent with that of Argoverse and nuScenes. The data extracted from the simulator includes: vehicle trajectories, 2D birds-eye-view map images, lane center-lines, and LiDAR point-cloud data. The physics engine (Unreal Engine 4.19) provides limited capability for producing actual surface contour points, with the provided in-built CARLA ray-tracing LiDAR sensor. Therefore, we utilize the depth image sensor to extract accurate depth information.
We use four depth sensors, each with $90 \deg$ field-of-view to capture depth images at each time-step. Then points are uniformly sampled on each image plane according to the number of channels and rotation frequency configuration of selected LiDAR. These points are corresponded with the pixel depth value by 2D bi-linear interpolation and projected into 3D space. The points are assigned a distance and an azimuth angle and converted from spherical coordinate to Cartesian coordinate system. This simulates precise object-shape contoured LiDAR point-cloud. Additional map Lane-center information is acquired from the map way-points, by maintaining a curvature at turning intersections.

In this paper, only birds-eye-view maps are used. These maps are utilized to build the distance transform of these images, as mentioned in Sec. 4.3 of the main paper. The drivable and non-drivable area information is extracted, based on the semantic segmentation of objects provided by the C++ API. The API is extended to specifically capture only the drivable area, and we utilize the Python API to save as a binary road mask with respect to ego-vehicle in simulation-time. 

As the calibration parameters are not provided by the Unreal Engine physics kit, the intrinsic and extrinsic camera parameters and the distortion parameters are estimated by multiple camera view bundle adjustment. These parameters are used to transform segmentation images from a perspective view to an orthographic birds-eye-view. The maps are synchronized with the trajectory data 20 points in the past trajectory segment and 30 points in the future trajectory segments for all agents in a snapshot. In each map, we fix the number of agents to 5, including the ego-vehicle. Engineered data can be further generated on similar lines, for specialized edge-case handling of model. We will release our extended CARLA dataset which presents multiple data modalities, corresponding simulation and data extraction codes for our fellow researchers. 


\begin{figure}[t]
\centering
\centerline{\includegraphics[width=12cm]{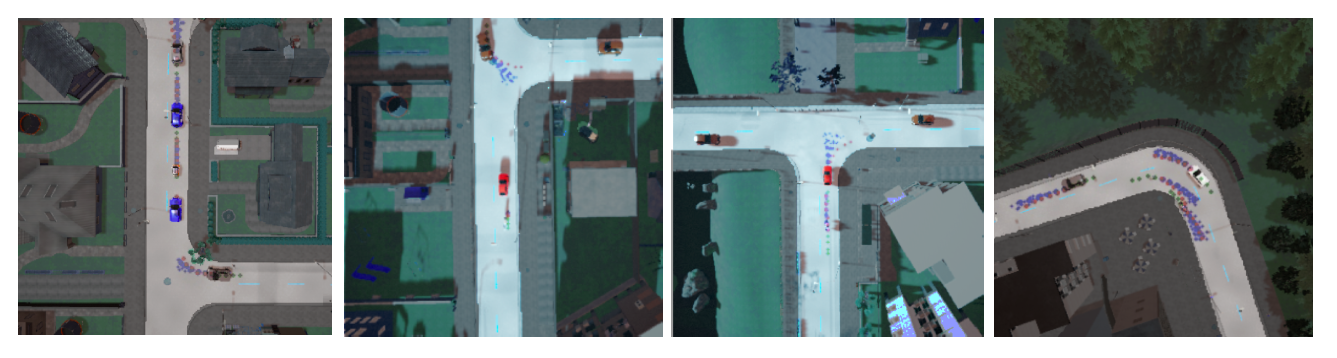}}
\caption{Visualizing predictions over simulated \textsc{CARLA} dataset at a snapshot of simulation-time with birds-eye-view image of drivable and non-drivable area and map. The \textsc{RED} points indicate ground truth, \textsc{GREEN} points indicate past trajectory points and \textsc{BLUE} indicates future trajectory points }
\label{fig:bev_map}
\end{figure}




\begin{table}
\caption{Results of baseline models and our proposed model in \textsc{CARLA} dataset. \textsc{Local-CAM-NF} is an ablation, whereas \textsc{AttGlobal-CAM-NF} is our full proposed model. The metrics are abbreviated as follows: \textsc{minADE}(\textbf{A}), \textsc{minFDE}(\textbf{B}),  \textsc{rF}(\textbf{C}), \textsc{DAO}(\textbf{D}), \textsc{DAC}(\textbf{E}). Improvements indicated by arrows.}

\setlength{\tabcolsep}{0.1pt}
\scriptsize
\centering
\begin{tabular}{p{2.9cm}p{1.2cm}p{1.2cm}p{1.2cm}p{1.2cm}p{0.75cm}}
\toprule
    \multirow{ 3}{*}{Model} & \multicolumn{5}{c}{\textsc{CARLA}} \\
    \cmidrule(r){2-6} & A~($\downarrow$) & B~($\downarrow$) & C~($\uparrow$)* & D~($\uparrow$)* & E~($\uparrow$)* \\
\midrule
        {\textsc{LSTM}}                             & 0.866 & 1.752 & 1.000 & 5.838 & 0.984 \\
        {\textsc{CSP}}  & 0.671 & 1.271 & 1.000 & 5.827 & 0.990 \\
        {\textsc{MATF-D}}      & 0.599 & 1.108 & 1.000 & 5.807 & \bf{0.992} \\
        {\textsc{DESIRE}}      & 0.748 & 1.281 & 1.745 & 21.68 & 0.969 \\
        {\textsc{MATF-GAN}}    & 0.566 & 1.015 & 1.212 & 11.94 & 0.988 \\
        {\textsc{R2P2-MA}} & 0.658 & 1.116 & 1.892 & 33.93 & 0.990 \\

    \midrule
        {\textsc{CAM}}                              & 0.695 & 1.421 & 1.000 & 5.615 & 0.981 \\
        {\textsc{CAM-NF}}                           & 0.543 & 1.006 & 2.135 & \bf{36.60} & 0.979 \\
        {\textsc{Local-CAM-NF}}                     & 0.483 & 0.874 & \bf{2.248} & 33.94 & 0.986 \\
        {\textsc{Global-CAM-NF}}                  & 0.490 & 0.898 & 2.146 & 33.00 & 0.986  \\
        {\textsc{AttGlobal-CAM-NF}}                 & \bf{0.462} & \bf{0.860} & 2.052 & 30.13 & 0.987 \\
\midrule
\end{tabular}
\label{tab:main_experiment}
\centering
\end{table}

\label{exp:performance_analysis}
\noindent \subsubsection{Performance Analysis:} 

Model generalizability depends on both the map characteristics as well as the statistical distribution of trajectory samples provided by the dataset. The CARLA dataset has simpler cases compared to the the ones in Argoverse motion forecasting. But there are scenic disturbances, like parked and moving agents in the private property-areas, which is considered as non-drivable area. Even considering these cases, the model is able to predict valid admissible trajectories, only influenced by the provided drivable area. We rigorously analyze the performance of our models in the CARLA dataset. The results are illustrated in Table~\ref{tab:main_experiment} where our model outperforms in \textsc{minADE} and \textsc{minFDE} metrics by about $15\%$ compared to baselines. \textsc{CAM-NF} model provides good performance in diversity metrics \textsc{DAO} illustrating its capability in improving diversity by using the flow-based decoder. From the results, our proposed \textsc{AttGlobal-CAM-NF} model is better able to mimic the true distribution of the autopilot in the simulator, showing considerable improvement in forecasting, over the baselines. 





\begin{table}
\caption{Analysis on $\alpha$ on \textsc{nuScenes}. 
}
\centering
\scriptsize
\begin{tabular}{lcccccccccccc}
\toprule
Model & & \textsc{minADE} $(\downarrow)$ & & \textsc{minFDE} $(\downarrow)$ & & \textsc{rF} $(\uparrow)$ & & \textsc{DAO} $(\uparrow)$ & & \textsc{DAC} $(\uparrow)$ \\
\hline
\shortstack{\textsc{AttGlobal-CAM-NF}(0.25)} & & \shortstack{0.775} & & \shortstack{1.349} & & \shortstack{2.310} & & \shortstack{\textbf{23.68}} & & \shortstack{0.912} \\
\shortstack{\textsc{AttGlobal-CAM-NF}(0.5)} & & \shortstack{\textbf{0.639}} & & \shortstack{\textbf{1.171}} & & \shortstack{\textbf{2.558}} & & \shortstack{22.62} & & \shortstack{\textbf{0.918}} \\
\shortstack{\textsc{AttGlobal-CAM-NF}(0.75)} & & \shortstack{0.666} & & \shortstack{1.246} & & \shortstack{2.431} & & \shortstack{23.04} & & \shortstack{0.914} \\
\shortstack{\textsc{AttGlobal-CAM-NF}(1.0)} & & \shortstack{0.840} & & \shortstack{1.679} & & \shortstack{1.970} & & \shortstack{22.83} & & \shortstack{0.904} \\
\bottomrule 
\end{tabular}
\label{tab:alpha_analysis}
\end{table}

\subsection{Analysis of $\alpha$}
\label{app:sec:alpha_exp}
We experiment with various values of $\alpha$|a degradation coefficient of constant velocity. As shown in Table~\ref{tab:alpha_analysis}, models tend to perform better when trained with a certain range of $\alpha$ less than $1.0$, reflecting that degrading the assumption of constant velocity (when $\alpha=1.0$) prevents the model to avoid trivial solutions and encourages it to actively seek for necessary cues for predictions in non-constant velocity. Among values we experimented, we choose to set $\alpha=0.5$ because the largest \textsc{rF} and \textsc{DAC} are observed in this setting, while remaining competitive in \textsc{DAO}, as well as the best precision in terms of \textsc{minErrors} of the prediction.
This is a desirable property in our task of diverse and admissible trajectory forecasting. A possible future direction of our degradation approach is to replace $\alpha$ with a learnable parametric function of noise centered at $1.0$ that can perturb the physical model of the world, similarly discussed in \cite{fortunato2017noisy}. 

\section{Scene Context Processing}
\subsection{The Drivable-area Map and Approximating $p$}
\label{app:sec:da and p}
\begin{figure}[H]
\centering
\includegraphics[width=0.8\linewidth]{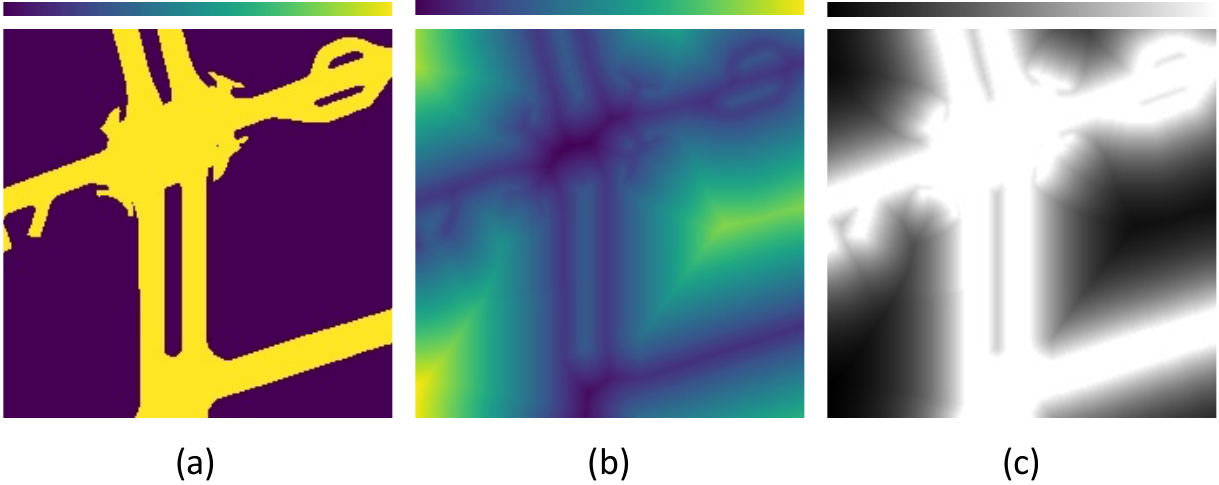}
\caption{(a) binary road mask, (b) the distance-transformed map, and (c) $\tilde{p}$ in nuScenes. The colorbars indicate the scales of the pixel value (increasing from left to right).
}
\label{fig:nusc_map_visualize}
\end{figure}

\begin{figure}[H]
\centering
\includegraphics[width=0.8\linewidth]{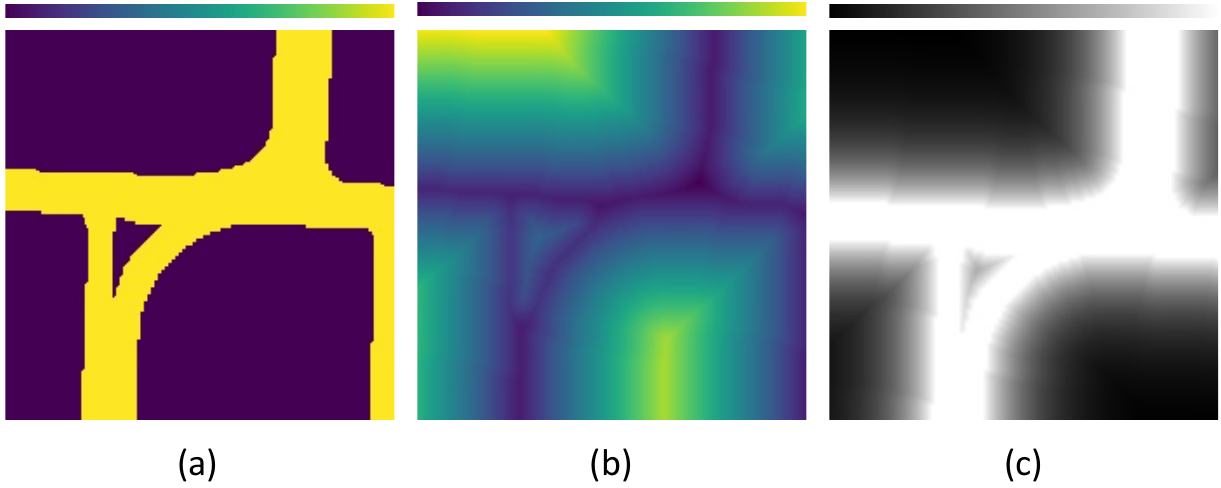}
\caption{(a) binary road mask, (b) the distance-transformed map, and (c) $\tilde{p}$ in Argoverse. The colorbars indicate the scales of the pixel value (increasing from left to right).
}
\label{fig:argo_map_visualize}
\end{figure}

\begin{figure}[H]
\centering
\includegraphics[width=0.8\linewidth]{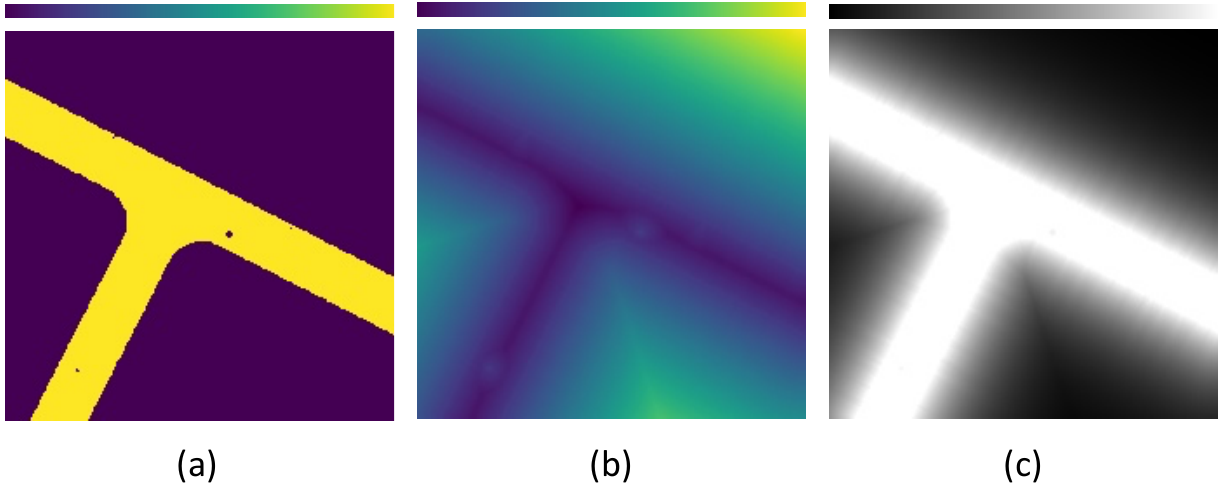}
\caption{(a) binary road mask, (b) the distance-transformed map, and (c) $\tilde{p}$ in CARLA. The colorbars indicate the scales of the pixel value (increasing from left to right).
}
\label{fig:carla_map_visualize}
\end{figure}

We utilize the $224 \times 224$ binary road masks in Sec.~A.1 and Sec.~A.2 to derive $\tilde{p}$, which is used in the evaluation of the reverse cross-entropy in our loss function at Eq.~(7), the main paper. Since $\tilde{p}$ is evaluated using $2\times2$ bilinear interpolation, the probability should be assigned to be gradually changing over the pixels, or the gradient would become $0$ almost everywhere. To this end, we apply the distance transform to the binary mask, as depicted in Fig~\ref{fig:nusc_map_visualize}(b),~Fig~\ref{fig:argo_map_visualize}(b),~Fig~\ref{fig:carla_map_visualize}(b). We utilize this transformed map for deriving $\tilde{p}$, as well as the scene context input $\bm{\Phi}$. As described in the main paper, our $\tilde{p}$ is defined based on the assumptions that every location on the drivable-area is equally probable for future trajectories to appear in and that the locations on non-drivable area are increasingly less probable, proportional to the distance from the drivable area. Hence, we first set the negative-valued pixels (drivable region) in a the distance-transformed map to have 0s assigned and subtract each pixel from the maximum so that the pixel values in the drivable-area are uniformly greatest over the map. Then, we normalize the map with the mean and standard deviation calculated over the training dataset to smooth the deviations. Finally, we apply softmax over the pixels to constitute the map as a probability distribution $\tilde{p}$, as depicted in Fig~\ref{fig:nusc_map_visualize}(c),~Fig~\ref{fig:argo_map_visualize}(c),~Fig~\ref{fig:carla_map_visualize}(c).

\subsection{Generating scene context input $\bm{\Phi}$}
\label{app:sec:scene_context_gen}
The scene context input $\bm{\Phi} \in \mathbb{R}^{224 \times 224 \times 3}$ is directly generated utilizing the the distance-transformed map. We augment 2 extra channels to the map, which embeds the unique pixel indices and Euclidean distances between each pixel and the center of the map. This way the ConvNet is enabled the access to the spatial position in the input thus it can extract the location-specific features from the scene context input. As similarly to the $\tilde{p}$, we normalize this augmented with the mean and standard deviation calculated over the training dataset to smooth the deviations.

\section{Additional Results Visualization}
\label{app:sec:visualizations}
\begin{figure}[H]
\centering
\includegraphics[width=\linewidth]{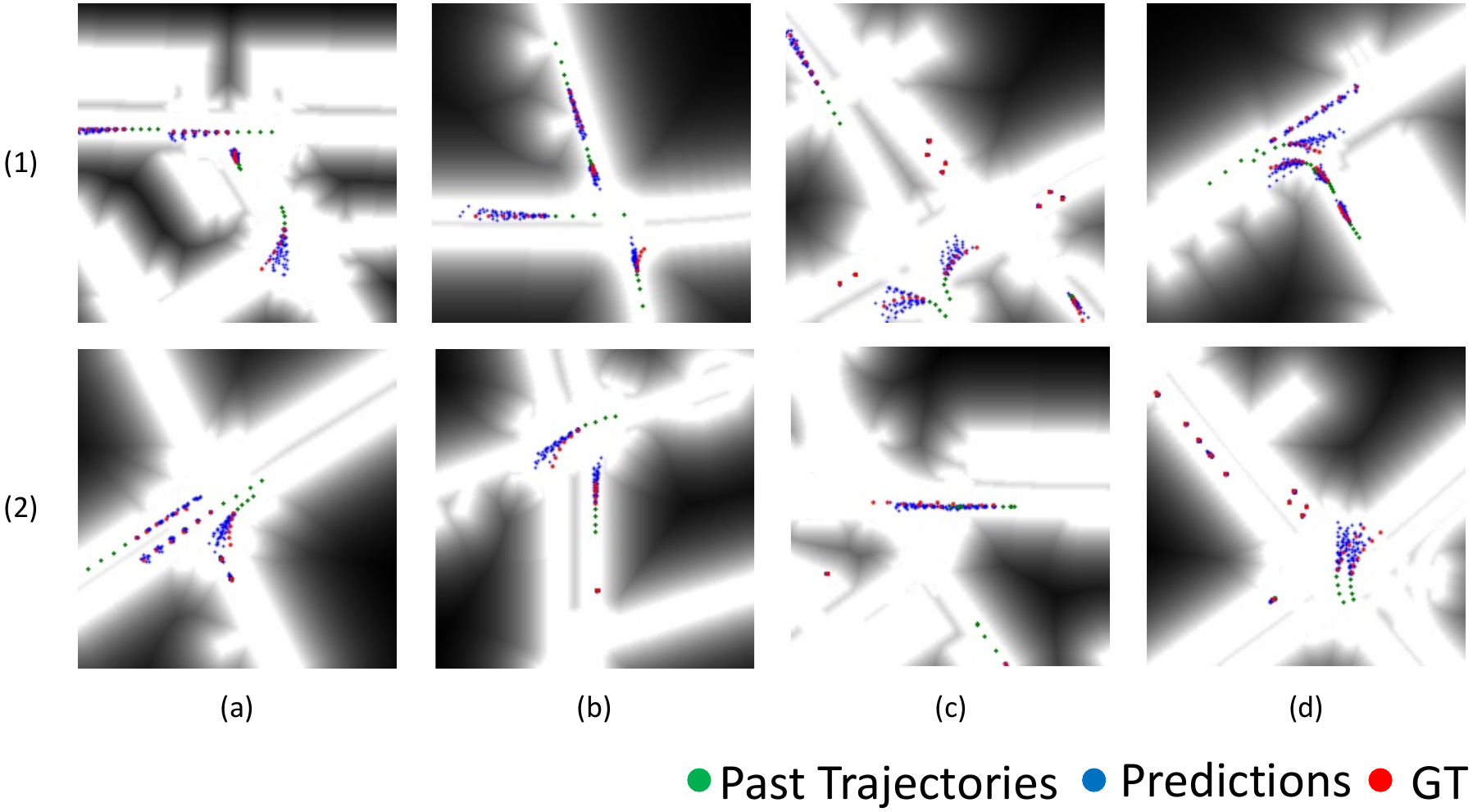}
\caption{Prediction results on nuScenes. Brighter background colors represent greater $~\tilde{p}$ values. Our approach predicts future trajectories that show diversity while remaining admissible.
}
\label{fig:nusc_results}
\end{figure}

\begin{figure}[H]
\centering
\includegraphics[width=\linewidth]{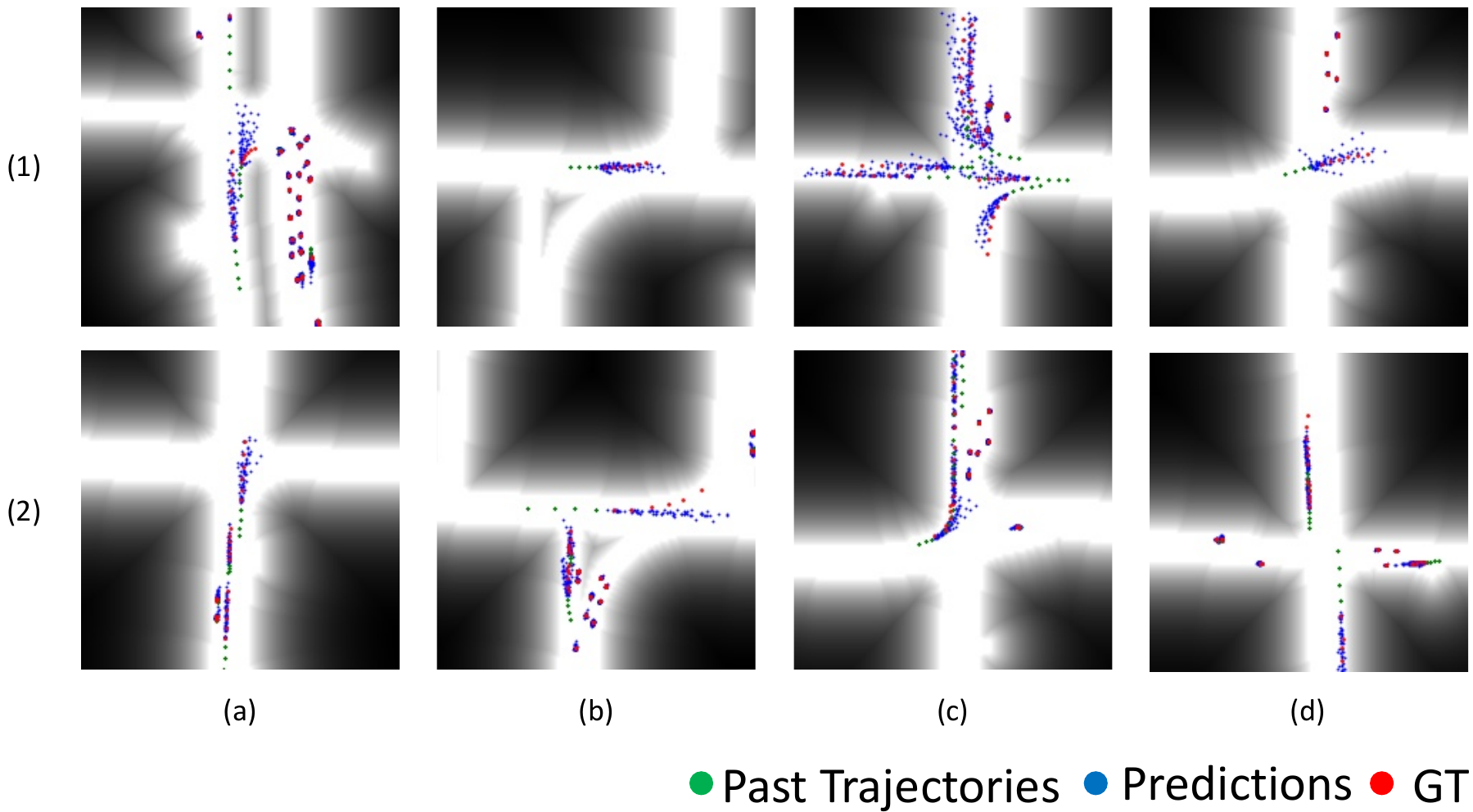}
\caption{Prediction results on Argoverse. Brighter background colors represent greater $~\tilde{p}$ values. Our approach predicts future trajectories that show diversity while remaining admissible.
}
\label{fig:argo_results}
\end{figure}

\begin{figure}[H]
\centering
\includegraphics[width=\linewidth]{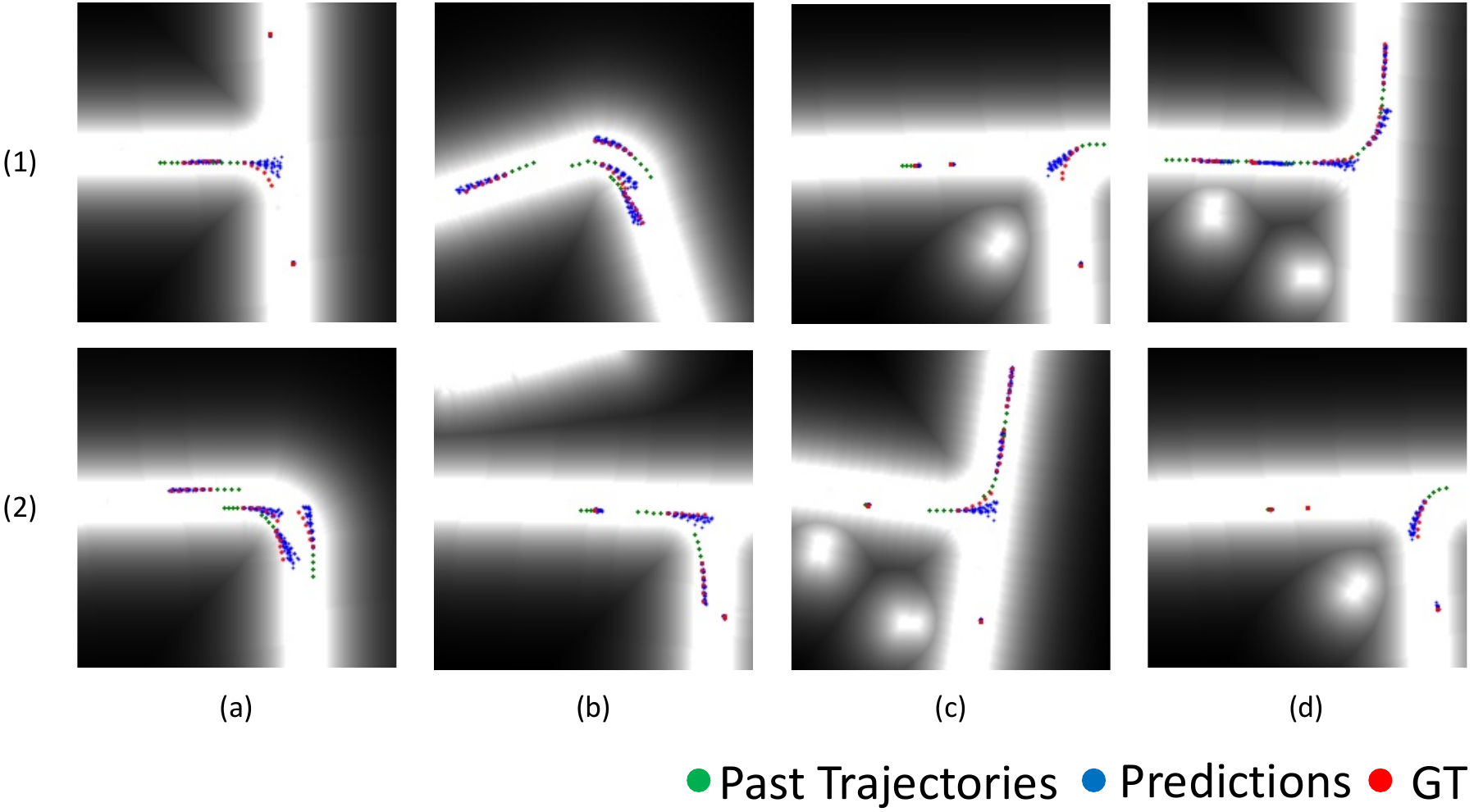}
\caption{Prediction results on CARLA. Brighter background colors represent greater $~\tilde{p}$ values. Our approach predicts future trajectories that show diversity while remaining admissible.
}
\label{fig:carla_results}
\end{figure}

Prediction results on nuScenes, Argoverse, and CARLA are illustrated in Fig~\ref{fig:nusc_results},
Fig~\ref{fig:argo_results}, and Fig~\ref{fig:carla_results}, respectively. In most cases throughout all three datasets, our model shows diverse and admissible predictions, even in the case of rapid left or right turns as in (2)(d) in Fig~\ref{fig:nusc_results} and (1)(d) and (2)(d) in Fig~\ref{fig:carla_results}. An clear insight on the quality of our model can be noticed when comparing predictions in cases of one path and multiple paths in front. For example, cases in (1)(b) and (2)(c) in  Fig~\ref{fig:nusc_results} show predictions are mostly going forward given the scene and the past trajectories going forward, whereas cases in (1)(c) and (2)(d) in Fig~\ref{fig:nusc_results} show predictions are diverse given the shape of the road and the past trajectories that start to make turns. One weakness of our experimental setting is that it is quite challenging for our model to predict an explicit left or right turn given the past trajectory going only forward as illustrated in the bottom-most trajectory in (1)(b) in Fig~\ref{fig:nusc_results} and (1)(a) in Fig~\ref{fig:carla_results}. One possible feature that can be added to avoid such cases would be to incorporate encoded lane information in our cross-agent interaction module along with trajectory encodings.


\section{Architecture and Training Details}
\label{app:sec:details}

\subsection{Training Setup}


For all experiments, we implemented models using the \texttt{PyTorch} 1.3.1 framework. We performed all data extraction and model training on a desktop machine with a NVIDIA TITAN X GPU card. We directly utilized the default implementations of Adam optimizer \cite{kingma2014adam} with the initial learning rate of $1.0\text{E-04}$ and an adaptive scheduler that halves the learning rate with patience parameter $3$ based on the sum of the \textsc{avgADE} and \textsc{avgFDE} in validation. The batch size is $64$ for all baselines and the proposed model except \textsc{AttGlobal-CAM-NF}, where the batch size of $4$ is used. We train the model for maximum 100 epochs, however, the early stopping is applied when the over-fitting of a model is observed.

\subsection{Hyperparameters}
Our experiments involve predicting 3 seconds of future trajectories of surrounding agents given a past record of 2 seconds, that includes the past trajectories and the scene context at the present time. The setting is common over all three datasets (nuScenes\cite{caesar2020nuscenes}, Argoverse\cite{chang2019argoverse}, and CARLA\cite{dosovitskiy1711carla}) that we experiment. The trajectories are represented as 2D position sequences recorded at every 0.5 seconds. The hyperparameters of the network structure is described in Table~\ref{tab:parameters} and Sec.~\ref{app:sec:network}.

\begin{table*}[!t]
\caption{Network Hyperparameters.
} 
\begin{center}
\scriptsize
\begin{tabular}{l c l c l c l}
\toprule
Operation & & Input (dim.) & & Output (dim.) & & Parameters\\ 
\hline
\multicolumn{7}{c}{\textsc{ConvNet}} \\
\multicolumn{7}{c}{\text{(Every Conv2D is with `same' padding, and followed by BN \cite{ioffe2015batch} \& ReLU.)}} \\
\hline
\textit{Conv2D} & & $\bm{\Phi}~(64, 64, 3)$ & & \textit{conv}1~(64, 64, 16) & & kernel$\coloneqq$(3,3) \\
\textit{Conv2D} & & \textit{conv}1 & & \textit{conv}2~(64, 64, 16) & & kernel$\coloneqq$(3,3) \\
\textit{MaxPool2D} & & \textit{conv}2 & & \textit{pool}2~(32, 32, 16) & & kernel$\coloneqq$(2,2), stride$\coloneqq$(2,2) \\
\textit{Conv2D} & & \textit{pool}2 & & \textit{conv}3~(32, 32, 32) & & kernel$\coloneqq$(5,5) \\
\textit{Conv2D} & & \textit{conv}3 & & \textit{conv}4~(32, 32, 6) & & kernel$\coloneqq$(1,1) \\
\textit{Dropout} & & \textit{conv}3 & & $\bm{\Gamma}_{g}$~(32, 32, 32) & & p$\coloneqq$0.5 \\
\textit{UpSample2D} & & \textit{conv}4 & & $\bm{\Gamma}_{l}$~(100, 100, 6) & & mode$\coloneqq$bilinear \\
\hline
\multicolumn{7}{c}{\textsc{Trajectory Encoding Module}} \\
\multicolumn{7}{c}{(\text{Repeated for} $a \in \{1,2,...,A\}$ \text{with variable encoding time length} $T_{\textit{past}}^{a}\coloneqq t^{a}_{s}+1$)} \\
\hline
\textit{Difference} & & $S^{a}_{\textit{past}}~(T^{a}, 2)$ & & $\textit{dS}^{a}_{\textit{past}}~(T^{a}, 2)$ & & \text{zero-pad (at the \textit{start time}}) \\
\textit{Fully-connected} & & $\textit{dS}^{a}_{\textit{past}}$ & & $\textit{tS}^{a}_{t}~(T^{a}, 128)$  & & \text{activation}$\coloneqq$\text{Linear} \\
\textit{LSTM} & & $\textit{tS}^{a}_{t}$ & & $h^{a}_{0}~(128)$ & & \text{zero initial states} \\
\hline
\multicolumn{7}{c}{\textsc{Cross-agent Interaction Module}} \\
\multicolumn{7}{c}{$\text{(Repeated for } a \in \{1,2,...,A\})$} \\
\hline
\textit{LayerNorm} & & $\textit{h}^{a}_{0}~(128)$ & & $\textit{h}_{ln}^{a}~(128)$ & & Layer Normalization \cite{ba2016layer} \\
\textit{Fully-connected} & & $\textit{h}_{ln}^{a} \forall a \in \{1,2,...,A\}$ & & $\bm{Q}~(A, 128)$  & & \text{activation}$\coloneqq$\text{Linear} \\
\textit{Fully-connected} & & $\textit{h}_{ln}^{a} \forall a \in \{1,2,...,A\}$ & & $\bm{K}~(A, 128)$ & & \text{activation}$\coloneqq$\text{Linear} \\
\textit{Fully-connected} & & $\textit{h}_{ln}^{a} \forall a \in \{1,2,...,A\}$ & & $\bm{V}$~(A, 128) & & \text{activation}$\coloneqq$\text{Linear} \\
\textit{Attention} & & $Q^{a}, \bm{K}, \bm{V}$ & & $\textit{h}_{\textit{atn}}^{a}~(128)$ & & \text{Scaled dot-product attention} \\
\textit{Addition} & & $h^{a}_{0}, \textit{h}_{\textit{atn}}^{a}$ & & $\tilde{h}^{a}$~(128) & & $\tilde{h}^{a} = h^{a}_{0} +  \textit{h}_{\textit{atn}}^{a}$\\
\hline
\multicolumn{7}{c}{\textsc{Local Scene Extractor}} \\
\multicolumn{7}{c}{$\text{(Repeated for } a \in \{1,2,...,A\})$} \\
\hline
\textit{Bilinear} & & $\bm{\Gamma}_l, \hat{S}^{a}_{t-1}~(2)$ & & $\gamma^{a}_{t}$~(6) & & \text{$2\times2$ Bilinear Interpolation}\\
\textit{Concatenate} & & $\tilde{h}^{a}, \gamma^{a}_{t}$ & & $\textit{hg}^{a}_{t}~(134)$ & & \text{-} \\
\textit{Fully-connected} & & $\textit{hg}^{a}_{t}$ & & $f\textit{hg}^{a}_{t}~(50)$ & & \text{activation}$\coloneqq$\text{Softplus} \\
\textit{Fully-connected} & & $f\textit{hg}^{a}_{t}$ & & $\textit{lc}^{a}_{t}~(50)$ & & \text{activation}$\coloneqq$\text{Softplus}\\
\hline
\multicolumn{7}{c}{\textsc{Agent-to-scene Interaction Module}} \\
\multicolumn{7}{c}{$\text{(Repeated for } a \in \{1,2,...,A\} \text{ with the fixed decoding time length } 6)$} \\
\hline
\textit{Flatten} & & $\hat{S}^{a}_{0:t-1}~(t,2)$ & & $\textit{f}\hat{S}^{a}_{0:t-1}~(12)$ & & \text{zero-pad (at the last), }$\hat{S}^{a}_{0} \coloneqq {S}^{a}_{0}$ \\
\textit{GRU} & & $\textit{f}\hat{S}^{a}_{0:t-1}$ & & $\hat{h}^{a}_{t}$ (150) & & \text{zero initial state} \\
\textit{Fully-connected} & & $\hat{h}^{a}_{t}$ & & $\textit{fh}^{a}_{t}$ (150) & & \text{activation}$\coloneqq$\text{Linear} \\
\textit{Fully-connected} & & $\bm{\Gamma}_{g}$ & & $\textit{f}\bm{\Gamma}$ (32, 32, 150) & & \text{activation}$\coloneqq$\text{Linear} \\
\textit{Attention} & & $\textit{fh}^{a}_{t}, \textit{f}\bm{\Gamma}$ & & $\alpha\bm{\Gamma}^{a}_{t}$ (32, 32, 1) & & \text{Additive attention} \\
\textit{Pool} & & $\bm{\Gamma}_{g}, \alpha\bm{\Gamma}^{a}_{t}$ & & $\tilde{\gamma}^{a}_{t}$ (32) & & $\text{Pixel-wise sum }(\bm{\Gamma}_{g} \odot \alpha\bm{\Gamma}^{a}_{t})$ \\
\textit{Concatenate} & & $\tilde{\gamma}^{a}_{t}, \hat{h}^{a}_{t}, \textit{lc}^{a}_{t}$ & & $\textit{gc}^{a}_{t}$ (232) & & \text{-} \\
\hline
\multicolumn{7}{c}{\textsc{Flow Decoder}} \\
\multicolumn{7}{c}{$\text{(Repeated for } a \in \{1,2,...,A\})$} \\
\hline
\textit{Fully-connected} & & $\textit{gc}^{a}_{t}$ & & $\textit{f}{gc}^{a}_{t}$ (50) & & \text{activation}$\coloneqq$\text{Softplus} \\
\textit{Fully-connected} & & $\textit{f}{gc}^{a}_{t}$ & & $\textit{ff}{gc}^{a}_{t}$ (50) & & \text{activation}$\coloneqq$\text{Tanh} \\
\textit{Fully-connected} & & $\textit{ff}{gc}^{a}_{t}$ & & $  \hat{u}^{a}_{t}~(2), \hat{\sigma}^{a}_{t}~(2,2)$ & & \text{activation}$\coloneqq$\text{Linear} \\
\textit{Expm} & & $\hat{\sigma}^{a}_{t}$ & & ${\sigma}^{a}_{t}~(2,2)$ & & impl. in \cite{bernstein1993some} \\
\textit{Constraint} & &
$\hat{\mu}^{a}_{t}, \hat{S}^{a}_{t-2:t-1}$
& &
$\mu^{a}_{t} (2)$
& &
$\alpha\coloneqq0.5$ \\
\textit{Random.} & &
\text{-}
& &
$z^{a}_{t} (2)$
& &
$z^{a}_{t} \sim \mathcal{N}(0, I)$ \\
\textit{Transform} & &
$z^{a}_{t}, \mu^{a}_{t}, \sigma^{a}_{t}$
& &
$S^{a}_{t} (2)$
& &
$S^{a}_{t} = \sigma^{a}_{t} z^{a}_{t} + \mu^{a}_{t}$ \\
\hline
\end{tabular}
\end{center}
\label{tab:parameters}
\end{table*}

\subsection{Network Details}
\label{app:sec:network}
In this section, we discuss the details on the network architecture of our model, which are not discussed in the main paper. The details of the the cross-agent attention and the agent-to-scene attention are particularly included in the discussion. We also give the edge case (e.g., bilinear interpolation at the position out of scene range) handling methods in our model. Refer to Table~\ref{tab:parameters} for the hyperparameters used in the components of our model.

We first discuss the details in the \textit{encoder}. As described in the main paper, the encoder is made of the trajectory encoding module and cross-agent interaction module. The trajectory encoding module utilizes a linear transform and a single layer LSTM to encode the observation trajectory for each agent $S^{a}_{\textit{past}} \equiv S^{a}_{t^{a}_{s}:0}$. To encode the observation trajectory $S^{a}_{\textit{past}}$, we first normalize it to $\textit{dS}^{a}_{\textit{past}}$ by taking the differences between the positions at each time step to have a translation-robust representation. Then, we linearly transform the normalized sequence $\textit{dS}^{a}_{t}$ to a high dimensional vector $\textit{tS}^{a}_{t}$ with Eq.~(\ref{eqn:encoder_linear_1}) then the vector is fed to the LSTM with Eq.~(\ref{eqn:encoder_lstm}) to update the state output $h^{a}_{t}$ and the cell state $c^{a}_{t}$ of the LSTM.

\begin{gather}
    \textit{tS}^{a}_{t} = \textsc{Linear}_1 (\textit{dS}^a_{t}) \label{eqn:encoder_linear_1} \\
    h^{a}_{t+1}, c^{a}_{t+1} = \textsc{LSTM}(\textit{tS}^{a}_{t}, h^{a}_{t}, c^{a}_{t}) \label{eqn:encoder_lstm}
\end{gather}

\noindent Wrapping Eq.~(\ref{eqn:encoder_linear_1}) and Eq.~(\ref{eqn:encoder_lstm}), along with the trajectory normalization process into a single function, we get Eq.~(1) in the main paper, where the same equation is copied as Eq.~(\ref{eqn:encoder_rnn}) for the convenience. Note that the initial states of the LSTM are set as zero-vectors.

\begin{gather}
h^{a}_{t+1} = \textsc{RNN}_{1} (S^a_{t}, h^{a}_{t})  \label{eqn:encoder_rnn}
\end{gather}


The cross-agent interaction module gets the set of agent embeddings $\bm{h}_0 = \{h^{1}_{0},...,h^{A}_{0}\}$ and models the agent-to-agent interaction via an attention architecture inspired by ``self-attention'' \cite{vaswani2017attention}. For each agent encoding $h^{a}_{0}$, we first apply \textit{Layer Normalization} \cite{ba2016layer} to get the normalized representation ${h}^{a}_{\textit{ln}}$ in Eq.~(\ref{eqn:encoder_layer_norm}) and calculate the query-key-value triple $(Q^{a},K^{a},V^{a})$ by linearly transforming ${h}^{a}_{\textit{ln}}$ with Eq.~(\ref{eqn:encoder_q_linear})-(\ref{eqn:encoder_v_linear}). Note that we empirically found that Layer Normalization gives a slight improvement to the model performance.

\begin{gather}
    {h}^{a}_{\textit{ln}} = \textsc{LayerNorm} (h^{a}_{0}) \label{eqn:encoder_layer_norm}
\end{gather}
\begin{gather}
    Q^{a} = \textsc{Linear}_{Q} ({h}^{a}_{\textit{ln}}) \label{eqn:encoder_q_linear}\\
    K^{a} = \textsc{Linear}_{K} ({h}^{a}_{\textit{ln}}) \label{eqn:encoder_k_linear}\\
    V^{a} = \textsc{Linear}_{V} ({h}^{a}_{\textit{ln}}) \label{eqn:encoder_v_linear}
\end{gather}

\noindent Next, we calculate the attention weights $\bm{w}^{a} \equiv \{w^{a}_{1}, w^{a}_{2},...,w^{a}_{A} \}$ by calculating \textit{scaled dot-product} between the query $Q^{a}$ and each key in the set of keys $\bm{K} \equiv \{K^{1},K^{2},...,K^{A} \}$ with Eq.~\ref{eqn:encoder_scaled_dot_product} and taking softmax over the set of products with Eq. \ref{eqn:encoder_softmax}.

\begin{gather}
    \textsc{SD}_{a, a'} = \frac{Q^{a} \cdot K^{a'}}{\sqrt{\textit{dim}({Q^{a})}}} \label{eqn:encoder_scaled_dot_product}\\
    \bm{w}^{a} = \textsc{Softmax}(\{\textsc{SD}_{a, 1},\textsc{SD}_{a, 2},...,\textsc{SD}_{a, A} \})\label{eqn:encoder_softmax}
\end{gather}

\noindent Finally, we get the interaction feature $\textsc{Attention}_{1} (Q^{a}, \bm{K}, \bm{V})$ in Eq.~(2) of the main paper, by taking the weighted average over $\bm{V} \equiv \{V^{1},V^{2},...,V^{A}\}$ using the weights $\bm{w}^{a}$ with Eq.~(\ref{eqn:encoder_weighted_average}).

\begin{gather}
    \textsc{Attention}_{1} (Q^{a}, \bm{K}, \bm{V}) = \sum_{a'=1}^{A} {w^{a}_{a'}  V^{a'}} = h^{a}_{\textit{atn}} \label{eqn:encoder_weighted_average}
\end{gather}

We now discuss the details in \textit{decoder}. As described in the main paper, the decoder is autoregressively defined and the previous outputs $\hat{S}^{a}_{0:t-1}$ are fed back. We encode the previous outputs using a GRU. Since the GRU requires a static shaped input at each time step, we reconfigure $\hat{S}^{a}_{0:t-1}$ by flattening and zero-padding it with Eq.~(\ref{eqn:decoder_reconfiguration}) and the reconfigured vector $\textit{f}\hat{S}^{a}_{0:t-1}$ is fed to the GRU at each decoding step with Eq.~(\ref{eqn:decoder_gru}).

\begin{gather}
    \textit{f}\hat{S}^{a}_{0:t-1} = \{\textsc{Flatten}(\hat{S}^{a}_{1:t-1}), 0,..., 0 \} \label{eqn:decoder_reconfiguration} \\
    \hat{h}^{a}_{t} = GRU(\textit{f}\hat{S}^{a}_{0:t-1}, \hat{h}^{a}_{t-1}) \label{eqn:decoder_gru}
\end{gather}

\noindent Wrapping Eq.~(\ref{eqn:decoder_reconfiguration}) and Eq.~(\ref{eqn:decoder_gru}) to a single function, we get Eq.~(3) in the main paper, where the same equation is copied as Eq.~\ref{eqn:decoder_rnn2} for the convenience. Note that the initial state of the GRU is set as a zero-vector.

\begin{gather}
    \hat{h}^{a}_{t} = \textsc{RNN}_{2} (\hat{S}^{a}_{1:t-1}, \hat{h}^{a}_{t-1}) \label{eqn:decoder_rnn2}
\end{gather}

\noindent The encoding $\hat{h}^{a}_{t}$ at each step is utilized in the agent-to-scene interaction module. The module is designed with an attention architecture inspired by ``visual-attention'' \cite{xu2015show}. We first linearly transform $\hat{h}^{a}_{t}$ and each pixel $\Gamma_{g}^{i}$ in the scene feature $\bm{\Gamma}_{g} \equiv \{\Gamma_{g}^{1},\Gamma_{g}^{2},...,\Gamma_{g}^{HW}\}$ with Eq~(\ref{eqn:decoder_u_linear}),(\ref{eqn:decoder_g_linear}).

\begin{gather}
    \textit{fh}_{t}^{a} = \textsc{Linear}_{\hat{h}} (\hat{h}^{a}_{t}) \label{eqn:decoder_u_linear}\\
    \textit{f}{\Gamma}^{i} = \textsc{Linear}_{{\Gamma_{g}}} (\Gamma_{g}^{i}) \label{eqn:decoder_g_linear}
\end{gather}

\noindent Then, we calculate the additive attention $\textsc{Attention}_{2} (\hat{h}^{a}_{t}, \bm{\Gamma})$ in Eq.~(4) of the main paper by taking the addition of $\textit{fh}_{t}^{a}$ to each $\textit{f}\Gamma^{i}$ followed by ReLU in Eq.~(\ref{eqn:decoder_addition_relu}), then applying softmax with Eq.~(\ref{eqn:decoder_softmax}).

\begin{gather}
    \textsc{SR}_{a,i} = \textsc{Relu} (\textit{fh}_{t}^{a} + \textit{f}\Gamma^{i}) \label{eqn:decoder_addition_relu}\\
    \textsc{Attention}_{2} (\hat{h}^{a}_{t}, \bm{\Gamma}) = \textsc{Softmax}(\{\textsc{SR}_{a, 1},\textsc{SR}_{a, 2},...,\textsc{SR}_{a, HW} \}) = \alpha\bm{\Gamma}^{a}_{t}  \label{eqn:decoder_softmax}
\end{gather}

\noindent Further details on our model's architecture and the hyperparameters, for instance the ConvNet and fully-connected layers discussed in the main paper are listed and defined in Table~\ref{tab:parameters}.

